\RequirePackage{fix-cm}
\documentclass[smallcondensed]{svjour3}     
\smartqed  
\usepackage{graphicx}
\usepackage{comment} 
\usepackage[colorinlistoftodos]{todonotes}
\usepackage{amsmath}
\usepackage{amssymb}
\usepackage{bm}
\usepackage{algorithm,algcompatible}
\usepackage{algpseudocode}
\usepackage{listings}
\usepackage[makeroom]{cancel}
\usepackage{dsfont}
\usepackage{float}
\usepackage{lscape}

\usepackage{array}
\usepackage{longtable}
\usepackage{caption}
\usepackage{xcolor,colortbl}
\newcommand\numberthis{\addtocounter{equation}{1}\tag{\theequation}}
\DeclareMathOperator*{\argmax}{\arg\!\max}
\definecolor{Gray}{gray}{0.85}
\algnewcommand\INPUT{\item[\textbf{Input:}]}%
\algnewcommand\OUTPUT{\item[\textbf{Output:}]}%
\newcommand{\Var}{\operatorname{Var}}
\makeatletter
\def\BState{\State\hskip-\ALG@thistlm}
\makeatother
\begin{document}

\title{A Hierarchical Bayesian Linear Regression Model with Local Features for Stochastic Dynamics Approximation 
}

\titlerunning{Bayesian Local Linear Regression}        

\author{Behnoosh Parsa         \and
		Keshav Rajasekaran		\and
        Franziska Meier			\and
        Ashis G. Banerjee 
}


\institute{B. Parsa \at
              Department of Mechanical Engineering, University of Washington, Seattle, WA, USA \\
              \email{behnoosh@uw.edu}           
           \and
           K. Rajasekaran \at
              Department of Mechanical Engineering, University of Maryland, College Park, MD, USA \\
            \email{keshavr@umd.edu}  
           \and
           F. Meier \at
              Autonomous Motion Department \\
              Max Planck Institute for Intelligent Systems, T\"{u}bingen, Germany \\
            \email{franzi.meier@gmail.com} 
           \and
           A. G. Banerjee \at
              Department of Industrial \& Systems Engneering and Department of Mechanical Engineering \\ University of Washington, Seattle, WA, USA \\
            \email{ashisb@uw.edu}  
}

\date{Received: date / Accepted: date}

\maketitle
\begin{abstract}
One of the challenges with model-based control of stochastic dynamical systems is that the state transition dynamics are involved, making it difficult and inefficient to make good-quality predictions of the states. Moreover, there are not many representational models for the majority of autonomous systems, as it is not easy to build a compact model that captures all the subtleties and uncertainties in the system dynamics. In this work, we present a hierarchical Bayesian linear regression model with local features to learn the dynamics of such systems. The model is hierarchical since we consider non-stationary priors for the model parameters which increases its flexibility. To solve the maximum likelihood (ML) estimation problem for this hierarchical model, we use the variational expectation maximization (EM) algorithm, and enhance the procedure by introducing hidden target variables. The algorithm is guaranteed to converge to the optimal log-likelihood values under certain reasonable assumptions. It also yields parsimonious model structures, and consistently provides fast and accurate predictions for all our examples, including two illustrative systems and a challenging micro-robotic system, involving large training and test sets. These results demonstrate the effectiveness of the method in approximating stochastic dynamics, which make it suitable for future use in a paradigm, such as model-based reinforcement learning, to compute optimal control policies in real time. 
\end{abstract}
\section{Introduction}
\label{sec:intro}
Analyzing the dynamics of nonlinear systems in various ways has been always an important area of research. This field includes nonlinear system identification \cite{ogunnaike1994process,nelles2013nonlinear}, modal analysis of dynamical system \cite{rudy2018data,tu2013dynamic}, system identification using neural networks \cite{narendra1990identification,hunt1992neural} and supervised learning of time series data, which is also known as function approximation or regression \cite{alaeddini2018linear,ting2011locally,vijayakumar2000locally,cleveland1988locally,junior2015regional}.

Most of the dynamical systems we want to understand and control are either stochastic, or they have some inherent noise components, which makes learning their models challenging. Moreover, in model-based control \cite{atkeson1998nonparametric,AndrewBagnell2014}, or optimal control \cite{ha2017multiscale}, often the model must be learned online as we get some measurement data throughout the process. Hence, the learning process not only has to result in accurate and precise approximations of the real-world behavior of the system but also must be fast enough so that it will not delay the control process. For example, in model-based reinforcement learning \cite{atkeson1998nonparametric,AndrewBagnell2014}, an agent first learns a model of the environment and then uses that model to decide which action is best to take next. If the computation of the transition dynamics takes very long, the predicted policy is no longer useful. On the other hand, if the control policy is learned based on an imprecise transition model, the policy cannot be used without significant modifications as shown in Atkeson (1994) \cite{atkeson1994using}. These learning procedures can be combined with Bayesian inference techniques to capture the stochasticities and/or uncertainties in the dynamical systems as well as the nonlinearities \cite{meier2014efficient,neal2012bayesian,kononenko1989bayesian}.

One of the popular methods to learn the transition dynamics is supervised learning. Usually, there are three different ways to construct the learning criteria: global, local, or a combination of both. From another perspective, these methods are classified into memory-based (lazy) or memory-less (eager) methods based on whether they use the training sets in the prediction process. An example of the former methodology is the $k$-nearest neighbor algorithm, and for the latter, the artificial neural network is a good representative. In an artificial neural network, the target function is approximated globally during training, implying that we do not need the training set to run an inference for a new query. Therefore, it requires much less memory than a lazy learning system. Moreover, the post-training queries have no effect on the learned model itself, and we get the same result every time for a given query. On the other hand, in lazy learning, the training set expands for any new query, and, thus, the model's prediction changes over time.
\subsection{Global Regression Methods for Learning Dynamics}
Most regression algorithms are global in that they minimize a global loss function. Gaussian processes (GPs) are popular global methods \cite{rasmussen2004gaussian} that allow us to estimate the hyperparameters under uncertainties. However, their computational cost ($\mathcal{O} (N^3)$ for $N$ observations) limits their applications in learning the transition dynamics for robot control. There have been efforts both in sparsifying Gaussian process regression (GPR) \cite{snelson2006sparse,titsias2009variational,quia2010sparse,matthews2016sparse,hensman2015mcmc} and developing online \cite{krizhevsky2012imagenet,wilson2016stochastic} and incremental \cite{gijsberts2013real,csato2002sparse} algorithms for the sparse GPR to make it applicable for robotic systems. Another challenging question is how to construct these processes, as for instance, how to initialize the hyperparameters or define a reasonable length scale. \cite{van2017convolutional} presents the convolutional Gaussian processes, in which a variational framework is adopted for approximation in GP models. The variational objective minimizes the KL divergence across the entire latent process \cite{matthews2016scalable}, which guarantees an exact model approximation given enough resources. The computational complexity of this algorithm is $ \mathcal{O} (NM^2)$ ($M \ll N$) through sparse approximation \cite{titsias2009variational}. Moreover, it optimizes a non-Gaussian likelihood function \cite{opper2009variational}. 
\subsection{Dynamics Learning with Spatially Localized Basis Functions} 
Receptive field-weighted regression (RFWR) in \cite{schaal1998constructive} is one of the prominent algorithms that motivates many incremental learning procedures. Receptive fields in nonparametric regression are constructed online and discarded right after prediction. Locally weighted regression (LWR), and a more advanced version of it, termed as locally weighted projection regression (LWPR) \cite{vijayakumar2000locally}, are two popular variants of RFWR, especially in control and robotics. The idea is that a local model is trained incrementally within each receptive field independent of the other receptive fields. Prediction is then made by blending the results of all the local models. The size and shape of the receptive fields and the bias on the relevance of the individual input dimensions are the parameters to be learned in these models. In addition, the learning algorithm allocates new receptive fields and prunes the extra ones as needed when new data are observed. The local models are usually simple like low-order polynomials \cite{hastie1993local}. These methods are robust to negative inference since the local models are trained independently. This contrasts with neural network models and is discussed in \cite{schaal1998constructive}. 

Another natural and interesting comparison is with the mixture of experts (ME) model \cite{jacobs1991adaptive,jordan1994hierarchical}. ME models are global learning systems where the experts compete globally to cover the training data, and they address the bias-variance problem \cite{fortmann2012understanding} by finding the right number of local experts and also the optimal distance metric for computing the weights in locally weighted regression \cite{atkeson1990using}. In \cite{xu1995alternative}, they define a mixture of Gaussian as the gating network and train both the gating network and the local models with a single-loop analytical version of the expectation-maximization (EM) algorithm. Schaal et al. \cite{schaal1998constructive} compare these two algorithms in \cite{schaal1998constructive}, and conclude that the performance of ME depends on the number of experts and their distributions in the input space. Moreover, it needs a larger training set than RFWR to provide comparably accurate results. The results of the two algorithms become indistinguishable when we increase the amount of training data and/or lower the noise. However, it is difficult to achieve these conditions since a large enough training set is not always available and the signal-to-noise ratio is not manually defined. Nevertheless, LWR and LWPR require many tuning parameters whose optimal values are highly data dependent. Therefore, we cannot always obtain robust and accurate results due to the highly localized training strategy of these methods, which does not allow the local models to benefit from the other models in their vicinities.

\subsection{Combining Local and Global 
Regression Methods for Dynamics Learning}
\begin{figure}[H]\centering 
\includegraphics[width=0.2\columnwidth]{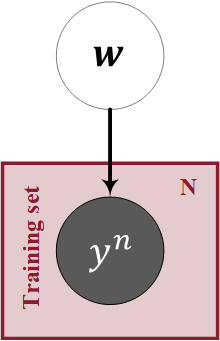}
\caption{Graphical model for a global linear regression problem solved using direct maximum likelihood (ML) estimation without any prior. $y^{(n)}$ is the observed random variable and $\mathbf{w}$ is the unknown model parameter. In this model, no prior knowledge is available about the unknown parameter.}
\label{fig:GraphicalModelGlobal} 
\end{figure}
Usually, the online performance of global methods depends on how well the distance metric is learned offline, which is based on how well the training data represent the input space. There are several concerns with linear models as discussed above. Therefore, one should develop better methods to tackle these problems. 

Local Gaussian regression (LGR) in \cite{meier2014efficient} is a probabilistic alternative to LWR, which transforms it into a localized inference procedure by employing variational approximation. This method combines the best of global and local regression frameworks by combining the well-known Bayesian regression framework with radial basis function (RBF) features for nonlinear function approximation. This is a top-down approach that takes advantage of the efficiency of the local regression and the accuracy of the global Bayesian regression and incorporates local features. In this method, maximizing the likelihood function is facilitated by the introduction of the hidden targets for every basis function, which act as links that connect observations to the unknown parameters via Bayes' law. In addition, the likelihood function is maximized using the Variational Expectation Maximization (EM) algorithm \cite{neal1998view,Jordan1999,Tzikas2008}. EM algorithm is a popular method that iteratively maximizes the likelihood function without explicitly computing it. The variational EM is an alternative algorithm that approximates the posterior distribution with a factorized function of the hidden variables in the model \cite{Tzikas2008}. The use of variational EM makes LGR a fast learning algorithm. A graphical model representation of this algorithm is shown in Fig. \ref{fig:GraphicalModelGlobal}.

Another useful characteristic of LGR is optimizing the RBF length scale so that the minimum number of local models is used for learning. This optimization reduces the size of the model significantly. In addition, it adopts the idea of \textit{Automatic Relevance Determination (ARD)} \cite{mackay1992bayesian,tipping2001sparse,neal2012bayesian} that results in a sparse model as described later in section \ref{sec:BLR}. Both these adjustments result in a compact model that does not occupy much memory as compared to the space needed to store the training set.

\subsection{Validity of the EM Algorithm}
The EM algorithm is a popular iterative method for solving maximum likelihood estimation problems, and its performance guarantees has been shown for the most generic EM algorithms \cite{balakrishnan2017statistical,wu1983convergence}, However,
there are very few theorems that prove its the general convergence properties of EM algorithms especially when applied to complex maximum likelihood problems. In problems with a bounded or convex loss functions, it is possible to find a generalization bound on the approximations made by the EM algorithm using Rademacher complexity \cite{cherkassky1999model,kakade2009complexity,grunwald2017tight,meir2003generalization}. This is a line of research that links the Bayesian and frequentist approaches together \cite{germain2016pac}. The idea is that the Bayes approach is powerful in constructing an estimator, whereas the frequentist approach provides a methodology for better evaluation. Therefore, by putting them together, one can come up with a more robust model design. However, these methods are limited to simple classes of problems \cite{opper1999general}, and are usually not applicable to hierarchical Bayesian models. This lack of applicability is primarily to the fact that the likelihood function cannot be described using a bounded or convex function. 

Moreover, most of the bounds found using the VC dimension or the Rademacher complexity are data-dependent, making them less popular for model (algorithm) evaluation. For instance, \cite{meir2003generalization} establishes data-dependent bounds for mixture algorithms with convex constraints and unbounded log-likelihood function. However, this is rarely the case in building complex models for high-dimensional datasets. 
Another effort has been to investigate whether the parameters updated by the EM algorithm converge to a stationary point in the log-likelihood function \cite{gunawardana2005convergence,gupta2011theory}. \cite{gunawardana2005convergence} restate the EM algorithm as a generalized alternating minimization procedure and used an information geometric framework for describing such algorithms and analyzing their convergence properties. They show that the optimal parameters found by the variational EM procedure is a stationary point in the likelihood function if and only if the parameters locally minimize the variational approximation error. This can only happen in two ways. The first way is if the variational error has stationary points in the likelihood function, which is guaranteed only if we know those stationary point prior to estimation. The other way is if the variational error is independent of the parameters, which is not possible if the variational family makes independence assumptions to ensure tractability. Therefore, such models have lower variational errors than those with independence assumptions \cite{gunawardana2005convergence}.

Dempster \textit{et al.} \cite{dempster1977maximum} prove a general convergence theorem for the EM algorithm. However, the proof is not constructed based on correct assumptions. \cite{wu1983convergence} and \cite{boyles1983convergence} point out flaws in the proof, and present a counter example for which the convergence of the generalized EM algorithm to the maxima of the likelihood function does not result in a single set of parameters. The convergence of these parameters are, therefore, highly dependent on the initial guess points and the properties of the likelihood function.

\subsection{Summary}
In this paper, we adopt the LGR model presented in \cite{meier2014efficient}, and modify it suitably for batch-wise learning of stochastic dynamics. We term this model the Batch-hierarchical Bayesian linear regression (Batch-HBLR) model. Moreover, we describe all the steps to derive the Bayes update equations for the posterior distributions of the model. We then analyze the convergence of the variational EM algorithm that is at the core of training the model. Subsequently, we evaluate its performance experimentally on three different dynamical systems, including a challenging external force-field actuated micro-robot. Results indicate good performance on all the systems in terms of approximating the dynamics closely with a parsimonious model structure leading to fast computation times during testing. We, therefore, anticipate our Batch-HBLR model to provide a foundation for online model-based reinforcement learning of robot motion control under uncertainty.   
\section{Hierarchical Bayesian Regression}
\label{sec:ProposedAlg} 
\subsection{Statistical Background}
\label{sec: Statistical Background}
Prior to formulating the regression model, it is useful to provide some statistical background. The multivariate normal (Gaussian) distribution for $\mathbf{x} \in \mathbb{R}^d$ and the univariate gamma distribution probability density functions (pdfs) are written in (\ref{eq:1}) and (\ref{eq:3}), respectively. 
\begin{equation}\label{eq:1}
    \begin{aligned}
    p(\mathbf{x}; \boldsymbol{\mu},\Sigma) = \frac{1}{\sqrt[]{(2\pi)^d|\boldsymbol{\Sigma}|}}\quad \exp\left( -\frac{1}{2}(\mathbf{x}-\boldsymbol{\mu})^\top\boldsymbol{\Sigma}^{-1}(\mathbf{x}-\boldsymbol{\mu})\right).
    \end{aligned}
\end{equation}

We use the following notation in the rest of the manuscript to refer to a normal distribution, $\mathcal{N}(\mathbf{x}; \boldsymbol{\mu},\boldsymbol{\Sigma})$, where $\boldsymbol{\mu}$ is the mean vector and $\boldsymbol{\Sigma}$ is the covariance matrix. The log-likelihood for $N$ i.i.d. samples drawn from $\mathcal{N}(\mathbf{x}; \boldsymbol{\mu},\boldsymbol{\Sigma})$ is:
\begin{equation}\label{eq:2}
    \begin{aligned}
    \log p(\mathbf{X}; \boldsymbol{\mu},\boldsymbol{\Sigma}) &= -\frac{N}{2}\log((2\pi)^d|\boldsymbol{\Sigma}|) -\frac{1}{2}\sum_{n=1}^N(\mathbf{x}_n-\boldsymbol{\mu})^\top\boldsymbol{\Sigma}^{-1}(\mathbf{x}_n-\boldsymbol{\mu}) +\textrm{const.}\\
    &\propto -\frac{N}{2}\log((2\pi)^d|\boldsymbol{\Sigma}|) -\frac{1}{2}\sum_{n=1}^N\textrm{Trace}\left(\boldsymbol{\Sigma}^{-1}(\mathbf{x}_n-\boldsymbol{\mu}) (\mathbf{x}_n-\boldsymbol{\mu})^\top\right)\\
    &= -\frac{N}{2}\log((2\pi)^d|\boldsymbol{\Sigma}|) -\frac{1}{2}\textrm{Trace}\left(\boldsymbol{\Sigma}^{-1}\sum_{n=1}^N(\mathbf{x}_n-\boldsymbol{\mu}) (\mathbf{x}_n-\boldsymbol{\mu})^\top\right).
    \end{aligned}
\end{equation}

Here, $\mathbf{X} \in \mathbb{R}^{N\times d}$ is a matrix containing all the samples, and $\mathbf{x}_n\in \mathbb{R}^d$ represents an individual sample. Moreover, in this manuscript, we use $\log$ for the natural logarithm.

For a random variable $x$ drawn from the Gamma distribution ($x \sim \mathcal{G}(\alpha,\beta)$), we use the following pdf,
\begin{equation}\label{eq:3}
    \begin{aligned}
    p(x; \alpha,\beta) = \frac{\beta^\alpha x^{\alpha-1}e^{-\beta x}}{\Gamma(\alpha)} \quad .
    \end{aligned}
\end{equation}
We use the following notation in the rest of the manuscript to refer to a gamma distribution, $\mathcal{G}(x; \alpha,\beta)$. Using Stirling's formula for the gamma function, we approximate $\log\Gamma(\alpha)$ for $\textit{Re}(\alpha)>0$ with $\alpha \log(\alpha)-\alpha$. Then, we get the following log-likelihood function for the gamma distribution,
\begin{equation}\label{eq:4}
    \begin{aligned}
    \log p(x; \alpha,\beta) &= \alpha \log(\beta) + (\alpha-1)\log(x) -\beta x-\alpha +\alpha \log(\alpha).
    \end{aligned}
\end{equation}

At this point, we also want to clarify the difference between the two notations $p(\mathbf{x};\boldsymbol{\theta})$ and $p(\mathbf{x} \mid \boldsymbol{\theta})$. We use the former when $\boldsymbol{\theta}$ is a vector of parameters, and the probability is a function of $\boldsymbol{\theta}$ and we call it the \textit{likelihood function}. In contrast, the latter denotes the conditional probability of $\mathbf{x}$ when $\boldsymbol{\theta}$ is a random variable.
\subsection{Bayesian Linear Regression Model with Local Features}
\label{sec:BLR}
In the Bayesian framework, consider the function $h(\mathbf{x}) \in \mathbb{R}$ and the variable $\mathbf{x} \in \Omega \subseteq \mathbb{R}^d$. We want to predict the function value $y^* = h(\mathbf{x}^*)$ at an arbitrary location $\mathbf{x}^* \in \Omega$, using a set of N noisy observations, $(\mathbf{X},\mathbf{Y})=\left\lbrace( \mathbf{x}^{(n)},y^{(n)})\right\rbrace_{n=1}^N$, where $y^{(n)} = h(\mathbf{x}^{(n)}) + \epsilon^{(n)}$. $\epsilon^{(n)}$ are independently drawn form a zero-mean Gaussian distribution,
\begin{equation}\label{eq:5}
p(\boldsymbol{\epsilon})= \mathcal{N}(\mathbf{0},\beta_y^{-1}\mathbf{I})
\end{equation}
where $\beta_y$ is the precision and $\boldsymbol{\epsilon} = \left[\epsilon^{(1)},\dots,\epsilon^{(N)}\right]^\top$.
Therefore, one can assume $(\mathbf{X},\mathbf{Y})$ is randomly sampled from a data generating distribution represented by $\mathcal{D}$, and denote $(\mathbf{X},\mathbf{Y}) \sim \mathcal{D}^N$ as the i.i.d. observations of $N$ elements, $\mathbf{x}^{(n)} \in \mathbb{R}^d$, and $y^{(n)} \in \mathbb{R}$. 

In basic (global) Bayesian regression, the function $h(\mathbf{x})$ is modeled as a linear combination of $P$ basis functions, and if these functions also include a bias term, then $P=d+1$.
\begin{equation}\label{eq:6}
h(\mathbf{x}^{(n)})= \mathbf{w}^\top \boldsymbol{\phi}(\mathbf{x}^{(n)})
\end{equation}
where, $\mathbf{w}=\left[w_1,\dots,w_P\right]^\top$ are the weights of the linear combination, and $ \boldsymbol{\phi}(\mathbf{x}^{(n)}) =\boldsymbol{\phi}^{(n)} = ({\mathbf{x}^{(n)}}^\top, 1)^\top$. By definition, $y\left(\mathbf{x}^{(n)}\right)= \mathbf{w}^\top \boldsymbol{\phi}\left(\mathbf{x}^{(n)}\right)+\epsilon^{(n)}$. Therefore, we write the likelihood function in the following form
\begin{equation}\label{eq:7}
p(\mathbf{Y}; \mathbf{w}) =\prod_{n=1}^N \mathcal{N}\left(y^{(n)}; \mathbf{w}^\top\boldsymbol{\phi}^{(n)},\beta_y^{-1}\right).
\end{equation}

In local regression, we introduce kernel functions, such as the Radial Basis Function (RBF), to emphasize the local observations about the point of interest. This emphasis results in more accurate predictions, while the computational complexity remains the same as the global regression problem. A better methodology is to use a Bayesian linear regression comprising local linear models, with a Gaussian likelihood for the regression parameters ($\tilde{\mathbf{w}}=[\mathbf{w}_1,\mathbf{w}_2,\dots,\mathbf{w}_M]$) written as:
\begin{equation}\label{eq:8}
p(\mathbf{Y}; \tilde{\mathbf{w}}) =\prod_{n=1}^N \mathcal{N}\left(y^{(n)}; \sum_{m=1}^M \mathbf{w}_m^\top\boldsymbol{\phi}_m^{(n)},\beta_y^{-1}\right),
\end{equation}
where $\mathbf{Y}=\left[y^{(1)},\dots,y^{(N)}\right]^\top \in \mathbb{R}^{N\times 1}$, and $\mathbf{w}_m \in \mathbb{R}^{P\times 1}$. $M$ is the dimensionality of the wighted feature vector $\boldsymbol{\phi}^{(n)}_m = \boldsymbol{\phi}_m(\mathbf{x}^{(n)})=\eta_m(\mathbf{x}^{(n)})\boldsymbol{\xi}_m \in  \mathbb{R}^{P}$, or the number of linear models. Here, we build the feature vector with linear bases plus a bias term, and define the $m^{th}$ feature vector as $\boldsymbol{\xi}_m=\left[(\mathbf{x}-\mathbf{c}_m)^\top,1 \right]^\top$. The weights for these $M$ spatially localized basis functions are defined by the Radial Basis Function (RBF) kernel, whereby the $m^{th}$ RBF feature weight is given by
\begin{equation}\label{eq:9}
\eta_m(\mathbf{x}) = \textrm{exp}\left[ -\frac{1}{2}(\mathbf{x}-\mathbf{c}_m)^\top\boldsymbol{\Lambda}_m^{-1}(\mathbf{x}-\mathbf{c}_m)\right].
\end{equation}
The RBF function is parameterized by its center $\mathbf{c}_m\in \mathbb{R}^d$ and a positive definite matrix $\boldsymbol{\Lambda}_m\in \mathbb{R}^{d\times d}$, both of which are estimated given the observations $\mathcal{O}= \left\lbrace\mathbf{x}^{(n)},y^{(n)}\right\rbrace_{n=1}^N$. Here, we consider the diagonal distance matrix $\boldsymbol{\Lambda}_m = \textrm{diag}\left(\lambda_{m_1}^{2},\dots, \lambda_{m_d}^{2}\right)$. 

\begin{figure}[H]\centering 
\includegraphics[width=0.5\columnwidth]{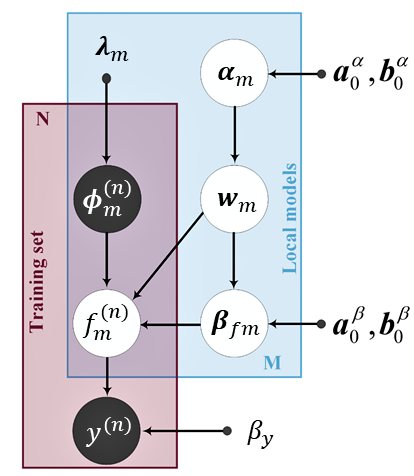}
\caption{Graphical model illustrating the algorithm (adapted from \cite{meier2014efficient}). The random variables inside the dark circles are observed. The variables inside the white circles are unknown. The variables represented with small black circles are the model parameters. This is a model with a hierarchical prior, since it includes priors on the parameters of the priors. That is why, we use a variational EM algorithm to solve the maximum likelihood estimation problem efficiently. The priors on the precision parameters are stationary, while those on the model weights are non-stationary.}
\label{fig:GraphicalModel}       
\end{figure}

In this methodology, we implicitly assume independence among the local models, as the correlations between the distant models are almost zero (due to the low value of the RBF). However, we have not yet used this assumption to reduce the computational cost of the EM algorithm (mentioned earlier in Section \ref{sec:intro} and described later in Section \ref{sec:inference}), to iteratively solve the maximum likelihood estimation problem for the models parameters. To do so, we introduce a hidden variable $f^{(n)}_m$ for every local model, and infer the parameters of each local model independently. Fig. \ref{fig:GraphicalModel} is a graphical model that illustrates how the hidden variables participate in both in the global and the local loops. Therefore, the complete likelihood of this graphical model is,
\begin{equation}\label{eq:10}
p(\mathbf{Y},\mathbf{f}; \boldsymbol{\theta}) =\prod_{n=1}^N \mathcal{N}\left(y^{(n)}; \mathds{1}^\top\tilde{\mathbf{f}}^{(n)},\beta_y^{-1}\right) \prod_{m=1}^M \mathcal{N}\left(f^{(n)}_m; \mathbf{w}_m^\top\boldsymbol{\phi}^{(n)}_m,\beta_{f_m}^{-1}\right).
\end{equation}
$\tilde{\mathbf{f}}^{(n)} = [f^{(n)}_1,f^{(n)}_2,\dots,f^{(n)}_M]^\top \in \mathbb{R}^{M}$, so $\mathbf{f}=[\tilde{\mathbf{f}}^{(1)},\tilde{\mathbf{f}}^{(2)},\dots,\tilde{\mathbf{f}}^{(N)}]^\top \in \mathbb{R}^{N\times M}$, and $\mathbf{Y} \in \mathbb{R}^{N}$. Later, in \textbf{Algorithm \ref{alg:main}}, we use $\mathbf{f}_m$ to denote the $m^{th}$ column of $\mathbf{f}$.
We place Gaussian priors on the regression parameters $\mathbf{w}_m$ (\ref{eq:11}) and gamma priors on the precision parameters $\{\beta_{fm},\boldsymbol{\alpha}_m\}$. It is worth mentioning that $\mathbf{A}_m$ is a diagonal matrix with the precision parameters along the main diagonal $\mathbf{A}_m =\textnormal{diag}\left( {\alpha}_m^{(1)},{\alpha}_m^{(2)},\dots,{\alpha}_m^{(P)}\right)$. When we write $\boldsymbol{\alpha}_m = \left[\boldsymbol{\alpha}_m^{(1)},\dots,\boldsymbol{\alpha}_m^{(P)}\right]^\top$, we refer to the vector that contains all the precision parameters of the local model $m$. Due to this diagonal structure of $\mathbf{A}_m$, the model learns the precision parameters for each feature independently. This property is very useful when adopting the idea of \textit{automatic relevance determination (ARD)} to automatically select the significant parameters by comparing the precision with a threshold value \cite{mackay1992bayesian,tipping2001sparse,neal2012bayesian}. ARD is an elegant method for sparsifying a learning model with many features. The prior on the weights (\ref{eq:11}) is a normal distribution; since, it is a conjugate prior, the posterior is also a normal distribution. 
\begin{equation}\label{eq:11}
p(\tilde{\mathbf{w}}; \boldsymbol{\alpha}) =\prod_{m=1}^M \mathcal{N}(\mathbf{w}_m;\mathbf{0}, {\mathbf{A}}_m^{-1}). \quad 
\end{equation}
After training, usually, some of these precision parameters converge to large numbers, which indicates that the corresponding feature does not play a significant role in representing the data. Thus, it would be reasonable to ignore those features in order to reduce the dimensionality of the regression model. This technique is called pruning, which largely alleviates the problem arising from the curse of dimensionality. 

The joint prior distribution of all the model parameters is:
\begin{equation}\label{eq:12}
\begin{aligned}
p(\tilde{\mathbf{w}},\boldsymbol{\beta}_{f},\mathbf{A}_m^{-1}) &=\prod_{m=1}^M p(\mathbf{w}_m \mid \boldsymbol{\alpha}_m)p(\beta_{f_m})\prod_{m=1}^M p(\boldsymbol{\alpha}_m)\\
&= \prod_{m=1}^M \mathcal{N}(\mathbf{w}_m; \mathbf{0}, \mathbf{A}_m^{-1})\mathcal{G}(\beta_{f_m}; a_0^\beta,b_0^\beta)\prod_{m=1}^M\prod_{p=1}^P\mathcal{G}(\boldsymbol{\alpha}_{m}^{(p)}; a_0^\alpha,b_0^{\alpha(p)}).
\end{aligned}
\end{equation}
To summarize, the model has hidden variables $\left\lbrace\mathbf{w}_m,\{f_m^{(n)}\}_{n=1}^N \right\rbrace_{m=1}^M$ and parameters $\boldsymbol{\theta} = \left\lbrace\beta_y,\{\beta_{fm},\boldsymbol{\alpha}_m, \lambda_{m}\}_{m=1}^M \right\rbrace$. 

\subsection{Variational Inference of Hidden Variables and Models Parameters}
\label{sec:inference}
We now want to infer the \textit{posterior probability} $p(\boldsymbol{\theta} \vert \mathbf{Y})$. One of the most popular methods to do so is maximum likelihood (ML). According to this approach, the ML estimate is obtained as 
\begin{equation}\label{eq:13}
\boldsymbol{\theta} = \argmax_{\boldsymbol{\theta}} p(\mathbf{Y};\boldsymbol{\theta}).
\end{equation}
As mentioned earlier, we use the EM algorithm to solve the ML estimation problem, and follow the exposition in \cite{Neal1998} and \cite{Bishop2006}. More specifically, we employ the variational EM framework as described below.

We rewrite the log-likelihood function as
\begin{equation}\label{eq:14}
\log p(\mathbf{Y};\boldsymbol{\theta}) = F(q,\boldsymbol{\theta})+\textmd{KL}(q\parallel p),
\end{equation}
with
\begin{equation}\label{eq:15}
F(q,\boldsymbol{\theta})=\int q(\mathbf{z})\log \left( \frac{p(\mathbf{Y},\mathbf{z}; \boldsymbol{\theta})}{q(\mathbf{z})}\right) d\mathbf{z},
\end{equation}
and
\begin{equation}\label{eq:16}
KL(q\parallel p)=-\int q(\mathbf{z})\log \left( \frac{p(\mathbf{z}\vert \mathbf{Y}; \boldsymbol{\theta})}{q(\mathbf{z})}\right) d\mathbf{z},
\end{equation}
where $q(\mathbf{z})$ is an arbitrary probability function for the hidden variable $\mathbf{z}$, and $\textmd{KL}(q\parallel p)$ is the Kullback-Leibler (KL) divergence between $p(\mathbf{Y}\vert\mathbf{z}; \boldsymbol{\theta})$ and $q(\mathbf{z})$. Since $\textmd{KL}(q\parallel p)\geq 0$, $\log p(\mathbf{Y};\boldsymbol{\theta}) \geq F(q,\boldsymbol{\theta})$. Hence, $F(q,\boldsymbol{\theta})$ is the \textit{lower bound} of the log-likelihood function. The EM algorithm maximizes the lower bound using a two step iterative procedure. Assume that the current value of the parameters is $\boldsymbol{\theta}_{old}$. The E-step maximizes the lower bound with respect to $q(\mathbf{z})$, which happens when $q(\mathbf{z})=p(\mathbf{z}\vert \mathbf{Y}; \boldsymbol{\theta})$. In this case, the lower bound is equal to the likelihood because $\textmd{KL}(q\parallel p)\geq 0$. In the M-step, we keep $q(\mathbf{z})$ constant, and maximize the lower bound with respect to the parameters to find a new value $\boldsymbol{\theta}_{new}$.

Now, if we substitute $q(\mathbf{z})=p(\mathbf{z}\vert \mathbf{Y}; \boldsymbol{\theta})$ in the lower bound and expand (\ref{eq:15}), we get
\begin{equation}\label{eq:17}
\begin{aligned}
F(q,\boldsymbol{\theta}) &= \int p(\mathbf{z}\vert \mathbf{Y}; \boldsymbol{\theta}_{old}) \log p(\mathbf{Y}, \mathbf{z}; \boldsymbol{\theta})d\mathbf{z} -  \int p(\mathbf{f}\vert \mathbf{Y}; \boldsymbol{\theta}_{old}) \log p(\mathbf{Y}, \mathbf{z}; \boldsymbol{\theta}_{old})d\mathbf{z}\\
&= Q(\boldsymbol{\theta},\boldsymbol{\theta}_{old})+\textit{constant},
\end{aligned}
\end{equation}
where the second term in the right-hand-side is the entropy of $p(\mathbf{z}\vert \mathbf{Y}; \boldsymbol{\theta}_{old})$ and does not depend on $\boldsymbol{\theta}$. The first term, which is usually named $Q(\boldsymbol{\theta},\boldsymbol{\theta}_{old})$, is the expectation of the likelihood of the \textit{complete data} ($\mathbb{E}_{p(\mathbf{z}\vert \mathbf{Y}; \boldsymbol{\theta}_{old})}\left[ \log p(\mathbf{Y}, \mathbf{z}; \boldsymbol{\theta}) \right]$). $Q(\boldsymbol{\theta},\boldsymbol{\theta}_{old})$ is commonly maximized in the M-step of the EM algorithm in the signal processing literature. To summarize, the two steps of the EM algorithm are,
\begin{align}
\textbf{E-step}: \textnormal{Compute  } p(\mathbf{z}\vert \mathbf{Y}; \boldsymbol{\theta}_{old})\label{eq:18}\\
\textbf{M-step}: \textnormal{Find  } \boldsymbol{\theta}_{new} = \argmax_{\boldsymbol{\theta}_{old}} Q(\boldsymbol{\theta},\boldsymbol{\theta}_{old}) .\label{eq:19}
\end{align}

In the variational EM algorithm, we approximate the posterior $p(\mathbf{z}\vert \mathbf{Y}; \boldsymbol{\theta}_{old})$ with a factorized function.
In this approximation, the hidden variable $\mathbf{z}$ is partitioned into $L$ partitions $z_i$ with $l=1,\dots,L$. $q(\mathbf{z})$ is also factorized in the same way $q(\mathbf{z})=\prod_{l=1}^{L} q_l(z_l)$. Given this assumption, the lower bound can be formulated as 
\begin{equation}\label{eq:20}
F(q,\boldsymbol{\theta}) = -\textmd{KL}(q_j\parallel\tilde{p})-\sum_{i\neq j}\int q_i \log q_i d\mathbf{z}
\end{equation}
where $q_j=q(\mathbf{z}_j)$, and 
\begin{equation}\label{eq:21}
\tilde{p}(\mathbf{Y},\mathbf{z}_j;\boldsymbol{\theta}) =\mathbb{E}_{i\neq j}\left[p(\mathbf{Y},\mathbf{z};\boldsymbol{\theta})\right] = \int p(\mathbf{Y},\mathbf{z}_j;\boldsymbol{\theta})\prod_{i\neq j}q_i d\mathbf{z}_i .
\end{equation}
The bound in (\ref{eq:20}) is maximized when the \textmd{KL} distance become zero, which is the case for $q_j(\mathbf{z}_j)=\tilde{p}(\mathbf{Y},\mathbf{z}_j;\boldsymbol{\theta})$. In other words, the optimal distribution is obtained from the following equation,
\begin{equation}
q^*_j(\mathbf{z}_j)= \mathbb{E}_{i\neq j}\left[p(\mathbf{Y},\mathbf{z};\boldsymbol{\theta})\right] + \textit{constant}\qquad \forall j=1,\dots,L.
\end{equation}
To summarize, the two steps of the variational EM algorithm are given by,
\begin{align}
\textbf{E-step}: \textnormal{Find } q_{new}(\mathbf{z}) = \argmax_{q(\mathbf{z})}  F(q,\boldsymbol{\theta}_{old}) \label{eq:23}\\
\textbf{M-step}: \textnormal{Find  } \boldsymbol{\theta}_{new} = \argmax_{\boldsymbol{\theta}} F(q_{new}(\mathbf{z}) ,\boldsymbol{\theta}) = \argmax_{\boldsymbol{\theta}} Q(\boldsymbol{\theta},\boldsymbol{\theta}_{old}).\label{eq:24}
\end{align}

\color{black}
These equations are the set of consistency conditions for the maximum of the lower bound for the factorized approximation of the posterior. They do not give us an explicit solution as they also depend on other factors. Hence, we need to iterate through all the factors and replace each of them one-by-one with its revised version. We now derive the updates for every factor of the hidden variables by considering the factorized posterior distribution,
\begin{equation}\label{eq:25}
\begin{aligned}
q(\mathbf{f},\tilde{\mathbf{w}},\tilde{\boldsymbol{\beta}}_{f},\tilde{\boldsymbol{\alpha}}) &=q(\mathbf{f})q(\tilde{\mathbf{w}})q(\tilde{\boldsymbol{\beta}}_{f})q(\tilde{\boldsymbol{\alpha}}).
\end{aligned}
\end{equation}
where $\tilde{\boldsymbol{\beta}}_{f} = \left[\beta_{f_1}, \dots,\beta_{f_M} \right]$ and $\tilde{\boldsymbol{\alpha}} = \left[\boldsymbol{\alpha}_{1}, \dots,\boldsymbol{\alpha}_{M} \right]$.

Note that assuming factorization between $\mathbf{f}$ and $\tilde{\mathbf{w}}$ automatically results in a factorized posterior distribution over all $\mathbf{f}_m$ and $\mathbf{w}_m$, and every element of $\boldsymbol{\alpha}_{m}$. To approximate the natural logarithm of the posterior distribution for the factors, we take the expected value of $q(\mathbf{f},\mathbf{w},\boldsymbol{\beta}_f,\boldsymbol{\alpha})$ over all the variables except the ones that we are solving for. In the resulting expression, we only keep the terms containing the variable of interest. By doing so, the variational Bayes update equations for the posterior mean and covariance of the parameters of each linear model are found in Equations (\ref{eq:26}) through (\ref{eq:43}). 

\footnotesize
\begin{equation}\label{eq:26}
\begin{aligned}
\log q(\mathbf{w}_m) &= \mathbb{E}_{\mathbf{f}_m,\mathbf{\beta}_{f_m},\alpha_m}\left[\sum_{n=1}^N\log\mathcal{N}(f^{(n)}_m ;\mathbf{w}_m^\top\boldsymbol{\phi}^{(n)}_m,\beta_{f_m}^{-1}) + \log\mathcal{N}(\mathbf{w}_m; \mathbf{0},{\mathbf{A}}_m^{-1})\right]\\
&=\mathbb{E}_{\mathbf{f}_m,\mathbf{\beta}_{f_m},\alpha_m} \left[\sum_{n=1}^N  \left[\log\beta_{f_m} -\frac{\beta_{f_m}}{2}\left(f_m^{(n)}-\mathbf{w}_m^\top\boldsymbol{\phi}_m^{(n)}\right)^\top\left(f_m^{(n)}-\mathbf{w}_m^\top\boldsymbol{\phi}_m^{(n)}\right)\right]\right.\\
&\left.\qquad\qquad\qquad\quad + \log(\vert\boldsymbol{A}_m\vert) -\frac{1}{2}\mathbf{w}_m^\top\boldsymbol{A}^{-1}_m\mathbf{w}_m \right]\\
&=\mathbb{E}_{\mathbf{f}_m,\mathbf{\beta}_{f_m},\alpha_m}  \left[-\frac{\beta_{f_m}}{2}\sum_{n=1}^N\left[{f_m^{(n)}}^2 + \phi_m^{(n)^\top}\mathbf{w}_m^\top\mathbf{w}_m\boldsymbol{\phi}_m^{(n)}   - 2f_m^{(n)}\mathbf{w}_m^\top\boldsymbol{\phi}_m^{(n)} \right]\right.\\
&\left.\qquad\qquad\qquad\quad-\frac{1}{2}\mathbf{w}_m^\top\boldsymbol{A}_m\mathbf{w}_m\right]\\
&= -\frac{1}{2}\mathbf{w}_m^\top\left( \mathbb{E}[\boldsymbol{A}_m]+\mathbb{E}[\beta_{f_m}] \sum_{n=1}^N{\boldsymbol{\phi}^{(n)}_m}{\boldsymbol{\phi}^{(n)}_m}^\top \right)\mathbf{w}_m + \mathbf{w}_m^\top\mathbb{E}[\beta_{f_m}]\sum_{n=1}^N\boldsymbol{\phi}^{(n)}_m\mathbb{E}[f^{(n)}_m]\\
&=\log\mathcal{N}(\mathbf{w}_m; \boldsymbol{\mu}_{w_m},\boldsymbol{\Sigma}_{w_m}).
\end{aligned}
\end{equation}
\normalsize
In the rest of the manuscript, we use ($\ \ \hat{}\ \ $) symbol to refer to the first moment of the approximation of the parameters. For instance, $\hat{\beta}_{f_m} = \mathbb{E}[\beta_{f_m}]$ and $\hat{\boldsymbol{A}}_m = \mathbb{E}[\boldsymbol{A}_m]$. Moreover, $\mathbb{E}[f^{(n)}_m] = \boldsymbol{\mu}^{\top}_{\mathbf{w}_m}\boldsymbol{\phi}^{(n)}_m$.

The posterior distribution of $\mathbf{w}_m$ is a Normal distribution; therefore, $\log q(\mathbf{w}_m)$ is a quadratic function of $\mathbf{w}_m$, which we refer as $\mathbf{J}(\mathbf{w}_m)$. From (\ref{eq:2}), we know that the negative inverse of the covariance matrix is equal to the second derivative of $\mathbf{J}(\mathbf{w}_m)$ with respect to $\mathbf{w}_m$.
The derivatives of the right hand side of (\ref{eq:11}) are:
\begin{equation}\label{eq:27}
\begin{aligned}
& \frac{\partial\mathbf{J}(\mathbf{w}_m)}{\partial \mathbf{w}_m} = -\left( \hat{\boldsymbol{A}}_m+\hat{\beta}_{f_m} \sum_{n=1}^N{\boldsymbol{\phi}^{(n)}_m}{\boldsymbol{\phi}^{(n)}_m}^\top \right)\mathbf{w}_m +\hat{\beta}_{f_m}\sum_{n=1}^N\boldsymbol{\phi}^{(n)}_m\mathbb{E}[f^{(n)}_m] \\
\end{aligned}
\end{equation}
\begin{equation}\label{eq:28}
\begin{aligned}
& \frac{\partial^2\mathbf{J}(\mathbf{w}_m)}{\partial \mathbf{w}_m^2} =  -\left( \hat{\boldsymbol{A}}_m+\hat{\beta}_{f_m} \sum_{n=1}^N{\boldsymbol{\phi}^{(n)}_m}{\boldsymbol{\phi}^{(n)}_m}^\top \right).\\
\end{aligned}
\end{equation}
Moreover, $\frac{\partial\mathbf{J}(\mathbf{w}_m)}{\partial \mathbf{w}_m}=0$ at the mean; hence, by setting (\ref{eq:12}) equal to zero and solving for $\mathbf{w}_m$, we get the mean of the posterior:
\begin{align}
& \boldsymbol{\Sigma}_{\mathbf{w}_{m}} = \left( \hat{\boldsymbol{A}}_m+\hat{\beta}_{f_m} \sum_{n=1}^N{\boldsymbol{\phi}^{(n)}_m}{\boldsymbol{\phi}^{(n)}_m}^\top\right)^{-1}\label{eq:29}\\
& \boldsymbol{\mu}_{\mathbf{w}_{m}}= \hat{\beta}_{f_m}\boldsymbol{\Sigma}_{{\mathbf{w}_{m}}}\sum_{n=1}^N\boldsymbol{\phi}^{(n)}_m\mathbb{E}[f^{(n)}_m] . \label{eq:30}
\end{align}
$\boldsymbol{\Sigma}_{\mathbf{w}_{m}} \in \mathbb{R}^{P\times P}$ is a diagonal matrix, and we refer to the elements on its main diagonal by $\sigma_{\mathbf{w}_{m}}^{(p)}$, where $p\in\left\{1,\dots,P\right\}$.

Similarly, we find the variational Bayes update for every element of the precision vector $\boldsymbol{\alpha}_m$. 
\footnotesize
\begin{equation}\label{eq:31}
\begin{aligned}
\log q(\boldsymbol{\alpha}_m^{(p)}) &= \mathbb{E}_{\mathbf{w}_{m}}\left[\log\mathcal{N}(\mathbf{w}_m^{(p)}; 0,\boldsymbol{\alpha}_m^{(p)}) + \log\mathcal{G}\left(\boldsymbol{\alpha}_m^{(p)}\mid a_0^{\alpha(p)},b_0^{\alpha(p)}\right)\right]\\
&= \mathbb{E}_{\mathbf{w}_{m}}\left[\log(\boldsymbol{\alpha}_m^{(p)})^{1/2}-\frac{\boldsymbol{\alpha}_m^{(p)}}{2}\left(\mathbf{w}_m^{(p)}\right)^2  +\left(a_0^{\alpha(p)}-1\right)\log(\boldsymbol{\alpha}_m^{(p)}) -b_0^{\alpha(p)}\boldsymbol{\alpha}_m^{(p)}\right]\\
&=\log\mathcal{G}\left(\boldsymbol{\alpha}_m^{\alpha(p)} \mid a_{Nm}^{\alpha(p)},b_{Nm}^{\alpha(p)}\right) .
\end{aligned}
\end{equation}
\normalsize
Hence,
\begin{align}
& a_{Nm}^{\alpha(p)} = a_0^{\alpha(p)} +\frac{1}{2}\label{eq:32} \\
& b_{Nm}^{\alpha(p)} = b_0^{\alpha(p)} +\frac{1}{2}\mathbb{E}\left[\left(\mathbf{w}_m^{(p)}\right)^2 \right]=
b_0^{\alpha(p)} + \frac{1}{2}\left(\left(\mu_{\mathbf{w}_m}^{(p)}\right)^2 + \sigma_{\mathbf{w}_{m}}^{(p)}\right).\label{eq:33}
\end{align}
We observe that $a_{Nm}^{\alpha(p)}$ is the same for all the models ($\forall m$) and every individual element ($p$) of the precision vector $\boldsymbol{\alpha}_m$. Therefore, we use $a_{N}^{\alpha}$ instead of $a_{Nm}^{\alpha(p)}$ later in \textbf{Algorithm \ref{alg:main}}.

Similarly, we derive the variational Bayes update for the posterior of the precision parameter $\boldsymbol{\beta}_f$ by computing
\footnotesize
\begin{equation}\label{eq:34}
\begin{aligned}
\log q(\boldsymbol{\beta}_{f_m}) &= \mathbb{E}_{\mathbf{w}_m,\mathbf{f}_{m}}\left[\sum_{n=1}^N\log\mathcal{N}\left(f^{(n)}_m; \mathbf{w}_m^\top\boldsymbol{\phi}^{(n)}_m,\beta_{f_m}^{-1}\right) + \log\mathcal{G}\left(\beta_{f_m} \mid a_0^\beta,b_0^\beta\right)\right]\\
&= \mathbb{E}_{\mathbf{w}_m,\mathbf{f}_{m}}\left[\log(\beta_{f_m})^{N/2}-\frac{\beta_{f_m}}{2}\sum_{n=1}^N\left[\left(f_m^{(n)}-\mathbf{w}_m^\top\boldsymbol{\phi}^{(n)}_m\right)^\top\left(f^{(n)}_m-\mathbf{w}_m^\top\boldsymbol{\phi}^{(n)}_m\right)\right] \right. \\ &\left. \quad\qquad\qquad+ \left(a_0^\beta-1\right)\log(\beta_{f_m}) -b_0^\beta\beta_{f_m}\right]\\
&=\log\mathcal{G}\left(\beta_{f_m} \mid a_{Nm}^\beta,b_{Nm}^\beta\right).
\end{aligned}
\end{equation}
\normalsize
The updates are
\begin{align}
& a_{Nm}^\beta = a_0^\beta +\frac{N}{2} \label{eq:35}\\
&\begin{aligned}
 b_{Nm}^\beta &= b_0^\beta +\frac{1}{2}\sum_{n=1}^N\left[\left(\mathbb{E}[f_m^{(n)}]-\mathbb{E}[\mathbf{w}_m]^\top\boldsymbol{\phi}^{(n)}_m\right)^\top\left(\mathbb{E}[f^{(n)}_m]-\mathbb{E}[\mathbf{w}_m]^\top\boldsymbol{\phi}^{(n)}_m\right)\right]\\
 &=b_0^\beta +\frac{1}{2}\sum_{n=1}^N\left[\left(\mu_{f_m^{(n)}}-\boldsymbol{\mu}_{\mathbf{w}_m}^\top\boldsymbol{\phi}^{(n)}_m\right)^\top\left(\mu_{f_m^{(n)}} -\boldsymbol{\mu}_{\mathbf{w}_m}^\top\boldsymbol{\phi}^{(n)}_m\right) + \sigma_{f_m} \right.\\
&\left. \qquad\qquad\quad \quad+ \textrm{Trace}\left( {\boldsymbol{\phi}^{(n)}_m}^\top\boldsymbol{\Sigma}_{\mathbf{w}_m}\boldsymbol{\phi}^{(n)}_m\right)\right].
\end{aligned}\label{eq:36}
\end{align}
In (\ref{eq:36}), $\sigma_{f_m}$ is the approximate variance of the $m^{th}$ local model. Again, since $a_{Nm}^\beta$ is identical for all the local models, we use $a_{N}^\beta$ instead.

Finally, the variational Bayes update equations for the posterior of mean and covariance of each local model target $\mathbf{f}_m = \left(f_m^{(1)},\dots, f_m^{(N)} \right)$ are found through the following steps:
\begin{equation}\label{eq:37}
\begin{aligned}
\log q(\tilde{\mathbf{f}}^{(n)}) &= \mathbb{E}_{\tilde{\mathbf{w}},\tilde{\boldsymbol{\beta}}_{f}}\left[\log\mathcal{N}\left(y^{(n)} \mid \mathds{1}^\top\tilde{\mathbf{f}}^{(n)},\beta_y^{-1}\right) + \log \mathcal{N}\left(\tilde{\mathbf{f}}^{(n)} \mid \tilde{\mathbf{F}}^{(n)},\mathbf{B}^{-1}\right)\right]\\
&= \mathbb{E}_{\tilde{\mathbf{w}},\tilde{\boldsymbol{\beta}}_{f}} \left[-\frac{\beta_y}{2}\left(y^{(n)}-\mathds{1}^\top\tilde{\mathbf{f}}^{(n)}\right)^\top\left(y^{(n)}-\mathds{1}^\top\tilde{\mathbf{f}}^{(n)}\right) -\frac{1}{2}\log(\mid\mathbf{B}\mid) \right.\\
&\left. \qquad\qquad\quad -\frac{1}{2}\left( \tilde{\mathbf{f}}^{(n)} - \tilde{\mathbf{F}}^{(n)}\right)^\top\mathbf{B}\left(\tilde{\mathbf{f}}^{(n)} - \tilde{\mathbf{F}}^{(n)}\right)\right]\\
&=\log\mathcal{N}(\tilde{\mathbf{f}}^{(n)}\mid \mathbf{\mu}_{\tilde{\mathbf{f}}^{(n)}},\mathbf{\Sigma}_{\tilde{\mathbf{f}}^{(n)}}),
\end{aligned}
\end{equation}
where $\tilde{\mathbf{F}}^{(n)} = \left[\mathbf{w}_1^\top\phi^{(n)}_1 ,\mathbf{w}_2^\top\phi^{(n)}_2,\dots,\mathbf{w}_m^\top\phi^{(n)}_m\right]^\top$, $\mathbf{B} =\left[\textrm{diag}(\beta_{f_1}, \dots,\beta_{f_m})\right]$, $\hat{\beta}_{f_m} = \mathbb{E}_{\beta_{f_m}}\left[\beta_{f_m}\right]= \frac{a^{\beta}_N}{b^{\beta}_{N,m}}$, and $\phi^{(n)}_m \in \mathbb{R}^{p}$.
\begin{equation}\label{eq:38}
\mathcal{N}(\tilde{\mathbf{f}}^{(n)}\mid \mathbf{\mu}_{\tilde{\mathbf{f}}^{(n)}},\mathbf{\Sigma}_{\tilde{\mathbf{f}}^{(n)}}) = \mathcal{N}(y^{(n)} \mid \mathds{1}^\top\tilde{\mathbf{f}}^{(n)},\beta_y^{-1}) \mathcal{N}(\tilde{\mathbf{f}}^{(n)} \mid \tilde{\mathbf{F}}^{(n)},\mathbf{B}^{-1}).
\end{equation}
We re-write the right hand side of (\ref{eq:38}) as an exponential function ($\exp(-J(\tilde{\mathbf{f}}^{(n)}))$), where $J(\tilde{\mathbf{f}}^{(n)})$ is a quadratic function of $\tilde{\mathbf{f}}^{(n)}$ and is defined as
\begin{equation}\label{eq:39}
\begin{aligned}
J(\tilde{\mathbf{f}}^{(n)}) &= \mathbb{E}_{\tilde{\mathbf{w}},\tilde{\boldsymbol{\beta}}_{f}} \left[\frac{\beta_y}{2}\left(\mathds{1}^\top \tilde{\mathbf{f}}^{(n)} - y^{(n)}\right)^\top\left(\mathds{1}^\top \tilde{\mathbf{f}}^{(n)} - y^{(n)}\right)\right.\\
&\left. \quad + \frac{1}{2}\left(\tilde{\mathbf{f}}^{(n)} - \tilde{\mathbf{F}}^{(n)}\right)^\top\mathbf{B}\left(\tilde{\mathbf{f}}^{(n)} - \tilde{\mathbf{F}}^{(n)}\right)\right].
\end{aligned}
\end{equation}
To find the parameters of $\mathcal{N}(\tilde{\mathbf{f}}^{(n)}\mid \mathbf{\mu}_{\tilde{\mathbf{f}}^{(n)}},\mathbf{\Sigma}_{\tilde{\mathbf{f}}^{(n)}})$, we compute the first and second derivatives of $J(\tilde{\mathbf{f}}^{(n)})$:
\begin{equation}\label{eq:40}
\frac{\partial J(\tilde{\mathbf{f}}^{(n)})}{\partial \tilde{\mathbf{f}}^{(n)}} = \beta_y \mathds{1}\left(\mathds{1}^\top \tilde{\mathbf{f}}^{(n)} - y^{(n)}\right) + \hat{\mathbf{B}}\left(\tilde{\mathbf{f}}^{(n)} - \mathbb{E}_{\mathbf{w}_m}\left[\tilde{\mathbf{F}}^{(n)}\right]\right)
\end{equation}
\begin{equation}\label{eq:41}
\frac{\partial^2 J(\tilde{\mathbf{f}}^{(n)})}{{\partial \tilde{\mathbf{f}}^{(n)}}^2} = \beta_y \mathds{1}\mathds{1}^\top + \hat{\mathbf{B}}.
\end{equation}
Hence, the covariance matrix $\mathbf{\Sigma}_{\tilde{\mathbf{f}}^{(n)}}$ is $\left( \beta_y \mathds{1}\mathds{1}^\top + \mathbf{B}\right)^{-1}$. Using the Sherman-Morrison formula, we reformulate the covariance matrix as
\begin{equation}\label{eq:42}
\mathbf{\Sigma}_{\tilde{\mathbf{f}}^{(n)}} = \hat{\mathbf{B}}^{-1} - \frac{\hat{\mathbf{B}}^{-1}\mathds{1}\mathds{1}^\top\mathbf{B}^{-1}}{\beta_y^{-1} + \mathds{1}^\top\hat{\mathbf{B}}^{-1}\mathds{1}}.
\end{equation}
Let us suppose $s=\beta_y^{-1} + \mathds{1}^\top\hat{\mathbf{B}}^{-1}\mathds{1}$. Then, the diagonal elements of $\mathbf{\Sigma}_{\tilde{\mathbf{f}}^{(n)}}$ are $\hat{\beta}_{f_m}^{-1} -\frac{\left(\hat{\beta}_{f_m}^{-1}\right)^2}{s}$. Note that the covariance matrix of the local models does not depend on the individual samples.

The minimum of $J(\tilde{\mathbf{f}}^{(n)})$ is attained when the first derivative is zero. Setting (\ref{eq:40}) to zero and solving for $\tilde{\mathbf{f}}^{(n)}$ gives us
\begin{equation}\label{eq:43}
\begin{aligned}
\mathbf{\mu}_{\tilde{\mathbf{f}}^{(n)}} &= \left(\beta_y\mathds{1}\mathds{1}^\top + \hat{\mathbf{B}}\right)^{-1}\left(\beta_yy^{(n)}\mathds{1} +\hat{\mathbf{B}}\mathbb{E}_{\mathbf{w}_m}\left[\tilde{\mathbf{F}}^{(n)}\right]\right)\\
&= \left(\hat{\mathbf{B}}^{-1} - \frac{\hat{\mathbf{B}}^{-1}\mathds{1}\mathds{1}^\top\mathbf{B}^{-1}}{\beta_y^{-1} + \mathds{1}^\top\hat{\mathbf{B}}^{-1}\mathds{1}}\right)\left(\beta_yy^{(n)}\mathds{1} +\hat{\mathbf{B}}\mathbb{E}_{\mathbf{w}_m}\left[\tilde{\mathbf{F}}^{(n)}\right]\right)\\
&= \mathbf{\Sigma}_{\tilde{\mathbf{f}}^{(n)}}\left(\beta_yy^{(n)}\mathds{1} +\hat{\mathbf{B}}\mathbb{E}_{\mathbf{w}_m}\left[\tilde{\mathbf{F}}^{(n)}\right]\right),
\end{aligned}
\end{equation}
where $\mathbb{E}_{\left\lbrace\beta_{f_m}\right\rbrace_{\forall m}}\left[\mathbf{B}\right]$ is written as $\hat{\mathbf{B}}$. The update for the individual target values is $\boldsymbol{\mu}_{\mathbf{f}_m} = \mathbf{f}_m+\frac{\left(\hat{\beta}_{f_m}^{-1}\right)^2}{s}\left(\mathbf{Y} - \mathbf{Y}_{pre}\right)$; $\mathbf{Y}_{pre} = \mathbb{E}_{\mathbf{w}_m}\left[\tilde{\mathbf{F}}^{(n)}\right]$.
\subsection{Length Scale Optimization}\label{sec: LengthScale}
To optimize the length scale $\boldsymbol{\lambda} =[\lambda_1,\dots, \lambda_M]$, we maximize the expected complete log likelihood under the variational bound
\begin{equation}\label{eq:44}
\begin{aligned}
\tilde{\lambda}^{opt} &= \argmax_{\boldsymbol{\lambda}} \mathbb{E}_{\mathbf{f},\tilde{\mathbf{w}},\tilde{\boldsymbol{\beta}}_{f},\tilde{\boldsymbol{\alpha}}}\log p(\mathbf{Y},\mathbf{f},\tilde{\mathbf{w}},\mathbf{\alpha},\tilde{\boldsymbol{\beta}}_{f} \mid \boldsymbol{\Phi},\boldsymbol{\lambda}).
\end{aligned}
\end{equation}
Here, by $\boldsymbol{\Phi}$, we refer to all the feature vectors $\boldsymbol{\phi}^{(n)}_m$ for every $n$ and $m$.
(\ref{eq:44}) nicely factorizes into independent maximization problems for each local model
\begin{equation}\label{eq:45}
\begin{aligned}
\lambda_m^{opt} &= \argmax_{\lambda_m} \mathbb{E}_{\mathbf{w}_m,\mathbf{f}_{m},\mathbf{\beta}_{f_m}}\sum_{n=1}^N\log\mathcal{N}\left(f^{(n)}_m \mid \mathbf{w}_m^\top\boldsymbol{\phi}^{(n)}_m,\beta_{f_m}^{-1}\right)
&=\argmax_{\lambda_m}K(\lambda_m), 
\end{aligned}
\end{equation}
which are optimized via gradient ascent. Simplifying (\ref{eq:45}), we have
\footnotesize
\begin{equation}\label{eq:46}
\begin{aligned}
\lambda_m^{opt} &= \argmax_{\lambda_m}\mathbb{E}_{\mathbf{w}_m,\mathbf{f}_{m}}\left[\log(\beta_{f_m})^{N/2}-\frac{\beta_{f_m}}{2}\sum_{n=1}^N\left[\left(f_m^{(n)}-\mathbf{w}_m^\top\boldsymbol{\phi}^{(n)}_m\right)^\top\left(f^{(n)}_m-\mathbf{w}_m^\top\boldsymbol{\phi}^{(n)}_m\right)\right]\right]\\
 &= \argmax_{\lambda_m}\mathbb{E}_{\mathbf{w}_m,\mathbf{f}_{m}}\left[-\frac{\beta_{f_m}}{2}\sum_{n=1}^N \left(f_m^{(n)}-\mathbf{w}_m^\top\boldsymbol{\phi}^{(n)}_m\right)^2\right]\\
&= \argmax_{\lambda_m}\left[-\frac{\beta_{f_m}}{2}\sum_{n=1}^N \left(\boldsymbol{\mu}_{f_m^{(n)}}-\boldsymbol{\mu}_{\mathbf{w}_m}^\top\boldsymbol{\phi}^{(n)}_m\right)^2\right].
\end{aligned}
\end{equation}
\normalsize
We evaluate the gradient at a single point each time and update the variables with the approximation of the gradient based on that single data point.
\begin{equation}\label{eq:47}
\begin{aligned}
\nabla_{\lambda_m}K &= \nabla_{\lambda_m} \sum_{n=1}^N\left[-\frac{\beta_{f_m}}{2}\left(\boldsymbol{\mu}_{f_m^{(n)}}-\boldsymbol{\mu}_{\mathbf{w}_m}^\top\boldsymbol{\phi}^{(n)}_m\right)^2\right]\\
&= -\sum_{n=1}^N\frac{\beta_{f_m}}{2}\left(\boldsymbol{\mu}_{f_m^{(n)}}-\boldsymbol{\mu}_{\mathbf{w}_m}^\top\boldsymbol{\phi}^{(n)}_m\right)\eta^{(n)}_m {\boldsymbol{\xi}^{(n)}_m}^{\top}\boldsymbol{\mu}_{\mathbf{w}_m}\textrm{power}(\mathbf{x}-\mathbf{c}_m,2)^{\top}\boldsymbol{\Lambda}^{-2},
\end{aligned}
\end{equation}
where $\textrm{power}(.,2)$ indicates the element-wise power operator. Also, $\nabla_{\mathbf{\lambda}_m}K \in \mathbb{R}^{d}$, and the update for $\mathbf{\lambda}_m$ is $\mathbf{\lambda}_m^{\textrm{new}} = \mathbf{\lambda}_m^{\textrm{old}} + \kappa\nabla_{\mathbf{\lambda}_m}$, where $\kappa$ is the learning rate. 

\subsection{Batch-wise Learning and Model Prediction}
\label{sec: Prediction}
The batch-wise learning of the regression model is summarized as \textbf{Algorithm \ref{alg:main}}, and is, henceforth, referred as the Batch-HBLR model. $w_{gen}$ is used as an activation threshold for the local models. If $x_n$ does not activate any of the available local models, the algorithm picks $\mathbf{x}_n$ as the center of a new local model. Moreover, we add a small positive number $\epsilon$ into the calculation of $\boldsymbol{\Sigma}_{\mathbf{w}_m}$ to avoid numerical issues during matrix inversion. The output is the set of local models learned for a batch of data. Note that all the update equations are local with the exception of $\boldsymbol{\mu}_{\mathbf{f}_m}$, the posterior mean of the hidden target. 

To find the predictive distribution, we marginalize the complete likelihood of the model first over the hidden variables $\mathbf{f}$
\begin{equation}\label{eq:48}
\begin{aligned}
\int \mathcal{N}\left(y^{*}; \mathds{1}^\top\tilde{\mathbf{f}}^{*},\beta_y^{-1}\right) \mathcal{N}\left(f^*; \mathbf{W}^\top\boldsymbol{\phi}^{*},\mathbf{B}^{-1}\right) df^* \\= \mathcal{N}\left(y^{*}; \sum_m^M \mathbf{w}_m^\top  \boldsymbol{\phi}^{*}_m ,\beta_y^{-1} + \mathds{1}^\top\mathbf{B}^{-1} \mathds{1}\right), 
\end{aligned}
\end{equation}
and then over the regression parameters $\mathbf{w}$
\begin{equation}\label{eq:49}
\begin{aligned}
\mathcal{N}\left(y^{*}; \sum_m^M \mathbf{w}_m^\top  \boldsymbol{\phi}^{*}_m ,\beta_y^{-1} + \mathds{1}^\top\mathbf{B}^{-1} \mathds{1}\right) \mathcal{N}\left(\mathbf{w}; \boldsymbol{\mu}_w,  \boldsymbol{\Sigma}_w  \right) d\mathbf{w}
\\=  \mathcal{N}\left(y^{*}; \sum_m^M \mathbf{w}_m^\top  \boldsymbol{\phi}^{*}_m ,\sigma^2(x^*)\right). 
\end{aligned}
\end{equation}
(\ref{eq:49}) represents the posterior with predictive mean $\sum_m^M \mathbf{w}_m^\top  \boldsymbol{\phi}^{*}_m $, which is the sum of all the local models predictions, and the predictive variance $\sigma^2(x^*) = \beta_y^{-1} + \mathds{1}^\top\mathbf{B}^{-1} \mathds{1} +   {\boldsymbol{\phi}^*_m}^\top \boldsymbol{\Sigma}_{\mathbf{w}_m} \boldsymbol{\phi}^*_m$.
\begin{algorithm}
\caption{Locally weighted Bayesian linear regression models implemented batch-wise} \label{alg:main}
\begin{algorithmic}[1]
\Procedure{Batch-HBLR}{$(\mathbf{X}, \mathbf{Y}) =\{\mathbf{x}^{(i)},y^{(i)}\}_{i=1}^N, T, \kappa, \epsilon, \delta, a_0^\alpha, b_0^\alpha, a_0^\beta, b_0^\beta, \beta_y, w_{gen}, \lambda_{init}$}\\

\State $LM \gets$ \Call{InitializeLM}{$\mathbf{X}, a_0^\alpha, b_0^\alpha, a_0^\beta, b_0^\beta, w_{gen}, \lambda_{init}$}

\vspace{1mm}
\State $a_{N}^\alpha = a_0^\alpha +\frac{1}{2}$ \Comment{$p$ is the number of LM parameters (equal to $d+1$, where $\mathbf{x} \in \mathbb{R}^d$)}
\State $a_{N}^\beta = a_0^\beta +\frac{N}{2}$ \\  
\State $t=0$
\While {$t<= T$} 
\State $t=t+1$
\For {$m = 1,\ldots,M$} 
%

\vspace{1mm}
\State $\mathbf{B} = \textrm{diag}(\hat{\beta}_{f_1}, \dots,\hat{\beta}_{f_m})$
\State $s=\beta_y^{-1} + \mathds{1}^\top\mathbf{B}^{-1}\mathds{1}$
\State ${\sigma}_{f_m} = \hat{\beta}_{f_m}^{-1} - \frac{\left(\hat{\beta}_{f_m}^{-1}\right)^2}{s}$ 
\\

\For {$n = 1, \ldots, N$}
\State $\eta_m(\mathbf{x}^{(n)}) = \textrm{exp}\left[ -\frac{1}{2}(\mathbf{x}^{(n)}-\mathbf{c}_m)^\top\boldsymbol{\Lambda}_m^{-1}(\mathbf{x}^{(n)}-\mathbf{c}_m)\right]$
\State $\boldsymbol{\phi}^{(n)}_m = \eta_m(\mathbf{x}^{(n)})\left[(\mathbf{x}^{(n)}-\mathbf{c}_m)^\top, 1 \right]^\top$  \Comment{$\boldsymbol{\phi}^{(n)}_m \in \mathbb{R}^{p}$}
\EndFor 

\vspace{2mm}
\State \Comment{Variational E-step to update the target variables and the moments of the posterior distributions of the regression models weights}
\vspace{1mm}
\State $\mathbb{E}[\mathbf{f}_m] = \left[\boldsymbol{\mu}^{\top}_{\mathbf{w}_m}\boldsymbol{\phi}^{(1)}_m,\dots,\boldsymbol{\mu}^\top_{\mathbf{w}_m}\boldsymbol{\phi}^{(N)}_m\right]^\top$ \Comment{$\mathbf{f}_m \in \mathbb{R}^{N}$ is the vector of the $m^{th}$ local model evaluations at the $N$ samples; we refer to the elements of this vector as $\mathbb{E}[f^{(n)}_m]$}

\vspace{1mm}
\State $\mathbf{Y}_{pre} = \left[\boldsymbol{\mu}^\top_{\mathbf{w}_m}\boldsymbol{\phi}^{(1)}_m,\dots,\boldsymbol{\mu}^\top_{\mathbf{w}_m}\boldsymbol{\phi}^{(N)}_m\right]^\top$ 

\State $\boldsymbol{\mu}_{\mathbf{f}_m} = \mathbf{f}_m+\frac{\left(\hat{\beta}_{f_m}^{-1}\right)^2}{s}\left(\mathbf{Y} - \mathbf{Y}_{pre}\right)$ \Comment{$\boldsymbol{\mu}_{\mathbf{f}_m} \in \mathbb{R}^N$}
\State $\hat{\mathbf{A}}_m = \textrm{diag}\left( \hat{\boldsymbol{\alpha}}^{(1)}_{1},\ldots,\hat{\boldsymbol{\alpha}}^{(p)}_{m} \right)$ \Comment{$\hat{\mathbf{A}}_m \in \mathbb{R}^{p \times p}$} 
\State $\boldsymbol{\Sigma}_{\mathbf{w}_m} = \left( (1+\epsilon)\hat{\mathbf{A}}_m +\hat{\beta}_{f_m}\sum_{n=1}^N{\boldsymbol{\phi}^{(n)}_m}{\boldsymbol{\phi}^{(n)}_m}^\top \right)^{-1}$ 
\Comment{$\boldsymbol{\Sigma}_{\mathbf{w}_m} \in \mathbb{R}^{p \times p}$}

\State $\boldsymbol{\mu}_{\mathbf{w}_m} = \hat{\beta}_{fm}\boldsymbol{\Sigma}_{\mathbf{w}_m}\sum_{n=1}^N{\boldsymbol{\phi}^{(n)}_m}\mathbb{E}[f^{(n)}_m]$
\Comment{$\boldsymbol{\mu}_{\mathbf{w}_m} \in \mathbb{R}^p$}

\vspace{2mm}
\State \Comment{Variational M-step to update the regression models hyperparameters}
\vspace{1mm}
\State $\begin{aligned}
b_{Nm}^\beta = &\ \ b_0^\beta +\frac{1}{2}\sum_{n=1}^N\left[\left(\mu_{f_m^{(n)}}-\boldsymbol{\mu}_{\mathbf{w}_m}^\top\boldsymbol{\phi}^{(n)}_m\right)^\top\left(\mu_{f_m^{(n)}} -\boldsymbol{\mu}_{\mathbf{w}_m}^\top\boldsymbol{\phi}^{(n)}_m\right) + \sigma_{f_m} \right.\\ & \left.+\textrm{Trace}\left( {\boldsymbol{\phi}^{(n)}_m}^\top\boldsymbol{\Sigma}_{\mathbf{w}_m}\boldsymbol{\phi}^{(n)}_m\right)\right]
\end{aligned}$

\State $\mathbf{b}_{Nm}^{\alpha(p)} = b_0^{\alpha(p)} + \frac{1}{2}\left(\left(\mu_{\mathbf{w}_m}^{(p)}\right)^2 + \sigma_{\mathbf{w}_{m}}^{(p)}\right)$ \\
\State $\hat{\boldsymbol{\alpha}}_{m}=\left[ \frac{a_{N}^\alpha}{\mathbf{b}_{Nm}^{\alpha(1)}}, \ldots, \frac{a_{N}^\alpha}{\mathbf{b}_{Nm}^{\alpha(p)}} \right]^\top$ \Comment{$\hat{\boldsymbol{\alpha}}_m \in \mathbb{R}^p$}
\State $\hat{\beta}_{f_m}=\frac{a_{N}^\beta}{b_{Nm}^\beta}$\Comment{$\hat{\beta}_{fm} \in \mathbb{R}$}

\vspace{2mm}
\State Update $\lambda_m$ using gradient ascent with $\kappa$ as the learning rate
\State $LM_m = \left\lbrace\mathbf{c}_m, \lambda_m, \hat{\beta}_{f_m}, \hat{\boldsymbol{\alpha}}_{m}, \boldsymbol{\mu}_{\mathbf{w}_m}, \boldsymbol{\Sigma}_{\mathbf{w}_m} \right\rbrace$
\EndFor
\State $\mathbf{Y}_{pre} = \left[\boldsymbol{\mu}^\top_{\mathbf{w}_m}\boldsymbol{\phi}^{(1)}_m,\dots,\boldsymbol{\mu}^\top_{\mathbf{w}_m}\boldsymbol{\phi}^{(N)}_m\right]^\top$ 
\State $nMSE[t] = \frac{\parallel\mathbf{Y}-\mathbf{Y}_{pre}\parallel_2^2}{\Var(\mathbf{Y})}$
\If {$\parallel nMSE[t]	- nMSE[t-1] \parallel	<= \delta$} \textrm{break} \Comment{Iteration threshold}
\EndIf
\EndWhile\\
\State \Return $LM$
\EndProcedure 
\algstore{myalg}
\end{algorithmic}
\end{algorithm}
\begin{algorithm}
\begin{algorithmic}[1]
\algrestore{myalg}
\Function{InitializeLM}{$\mathbf{X}, a_0^\alpha, b_0^\alpha, a_0^\beta, b_0^\beta, w_{gen}, \lambda_{init}$}\label{function}\\
\State $M = 1$ \Comment{$M$ is the index of the local models}
\State $LM_1 = \left\lbrace\mathbf{c}_1=\mathbf{x}^{(1)}, \lambda_1 = \lambda_{init}, \hat{\beta}_{f_1} = \frac{a_0^{\beta}}{b_0^{\beta}}, \hat{\boldsymbol{\alpha}}_{1} = \frac{a_0^{\alpha}}{b_0^{\alpha}}\mathds{1}_p, \boldsymbol{\mu}_{\mathbf{w}_1} = \mathbf{0}_p, \boldsymbol{\Sigma}_{\mathbf{w}_1} = \mathbf{0}_{p \times p} \right\rbrace$ 
\For {$n = 1, \ldots, N$} 
\For {$m = 1, \ldots, M$} 
\State $\boldsymbol{\Lambda}_m=\lambda_{init}^2\mathbf{I}_d$ \Comment{$\mathbf{I}_d \in \mathbb{R}^{d\times d}$ is the identity matrix}
\State $\eta_m(\mathbf{x}^{(n)}) = \textrm{exp}\left[ -\frac{1}{2}(\mathbf{x}^{(n)}-\mathbf{c}_m)^\top\boldsymbol{\Lambda}_m^{-1}(\mathbf{x}^{(n)}-\mathbf{c}_m)\right]$
\EndFor\\
\If {$\eta_m \left( \mathbf{x}^{(n)} \right)< w_{gen}, \quad \forall m = 1, \ldots, M$} 
\State $M = M +1$ 
\State $LM_M =\begin{aligned} \left\lbrace\mathbf{c}_M = \mathbf{x}^{(n)}, \lambda_M = \lambda_{init}, \hat{\beta}_{f_M} = \frac{a_0^{\beta}}{b_0^{\beta}}, \hat{\boldsymbol{\alpha}}_{M} = \frac{a_0^{\alpha}}{b_0^{\alpha}}\mathds{1}_p, \boldsymbol{\mu}_{\mathbf{w}_m} = \mathbf{0}_p,\right.\\ \left. \boldsymbol{\Sigma}_{\mathbf{w}_m} = \mathbf{0}_{p \times p} \LARGE{ d}\right\rbrace\end{aligned}$
\EndIf 
\EndFor\\
\State \Return $LM$  
\EndFunction 
\end{algorithmic}
\end{algorithm}

For better prediction, we divide the training set into mini-batches when the training set is very large, and train the model for each mini-batch separately. To make the prediction for a new location $\mathbf{x}^*$, we find the correct mini-batch model(s) that $\mathbf{x}^*$ belongs to, and then predict $\mathbf{x}^*$ using (\ref{eq:49}). The prediction from each model is averaged for the points that lie within the intersection of two or more mini-batches. The procedure for preparing data and training the model on all the batches is described in \textbf{Algorithm \ref{alg:training}}.
The procedure for testing the models is stated in \textbf{Algorithm \ref{alg:testing}}.
\begin{algorithm}[!t]
\caption{Data preparation and model training} \label{alg:training}
\begin{algorithmic}[1]
\Procedure{Training}{$(\mathbf{X}, \mathbf{Y}) =\{\mathbf{x}^{(i)},y^{(i)}\}_{i=1}^N$}\\
\State \textrm{Find all the instances corresponding to the change in the control signal ($\mathbf{T}^*$)}\\
\State \textrm{If possible, evenly divide the training input space into S segment using $\mathbf{T}^*$. If not divide such that only the size of the last segment is different ($(\mathbf{X}, \mathbf{Y}) =\cup_{s=1}^{S}(\mathbf{X}_s, \mathbf{Y}_s)$)}
\If {$\Var(\mathbf{Y}_s)==0.0$}
\State $\boldsymbol{\epsilon} \sim \mathcal{N}(0.0,10^{-8})$
\State $\mathbf{Y}_s = \mathbf{Y}_s + \boldsymbol{\epsilon}$
\EndIf\\
\For {$\forall s$}
\State $LM_s=$\textit{Batch-HBLR}($\mathbf{X}_s, \mathbf{Y}_s$)
\EndFor
\State \Return $LM$
\EndProcedure 
\end{algorithmic}
\end{algorithm}
\begin{algorithm}[!t]
\caption{Model testing} \label{alg:testing}
\begin{algorithmic}[1]
\Procedure{Test}{$(\mathbf{X}, \mathbf{Y}) =\{\mathbf{x}^{(i)},y^{(i)}\}_{i=1}^N$}\\
\For {$i = 1,\dots, N$}
\State Determine the segments $\mathbf{x}^{(i)}$ belongs to (store their indices in $S$), and use the model learned for that segment ($LM_s$) to make prediction (Eq. \ref{eq:34})\\
\If {there is only one segment to which $\mathbf{x}^{(i)}$ belongs}
\State $\mathbf{Y}^{(i)}_{pre} =\sum_m^M \mathbf{w}_m^\top\boldsymbol{\phi}^{i}_m $ \\
\ElsIf {there are more segments}
\For {$s \in S$}
\State $\mathbf{y}(s) =\sum_m^M \mathbf{w}_m^\top\boldsymbol{\phi}^{i}_m $ \Comment{Using $LM_s$}
\EndFor \\
\State $\mathbf{Y}^{(i)}_{pre} =\frac{\sum_{s=1}^{\vert S\vert} \mathbf{y}(s)}{\vert S\vert} $ \Comment{$\vert S\vert$ is the number of active segments}
\EndIf\\
\EndFor\\
\State $nMSE = \frac{\parallel\mathbf{Y}-\mathbf{Y}_{pre}\parallel_2^2}{\Var(\mathbf{Y})}$ \\
\State \Return $nMSE$
\EndProcedure 
\end{algorithmic}
\end{algorithm}

\section{Theoretical Analysis}
\label{sec: Theoretical}
We have explained how to solve an MLE problem for the graphical model in Fig. \ref{fig:GraphicalModel} using a variation EM algorithm, and derived the corresponding update equations for the unknown model parameters in the previous Section. Therefore, now, an important question is whether the variational EM algorithm has the desired performance. As discussed in the Introduction, there have been many efforts toward finding convergence guarantees of various EM algorithms. Here, we show that, assuming a factorized posterior and considering the monotonicity of the EM algorithm as established in Theorem 2.1 of \cite{gupta2011theory}, the $(t+1)^{th}$ update is never less likely than the $t^{th}$ update. Alternatively, this means that improving the $Q$-function in (\ref{eq:24}) never makes the log-likelihood function worse. Let us restate the Theorem for a part of the Batch-HBLR model:

\begin{theorem}\label{thm:1}
Let random variables $X$ and $Y$ have parametric densities with parameter $\theta\in \Omega$. Suppose the support of $X$ does not depend on $\theta$, and the Markov relationship $\theta\rightarrow X \rightarrow Y$, that is,
\[p(y\vert x,\theta) = p(y\vert x) \]
holds for all $\theta\in \Omega$, $X\in\mathcal{X}$ and $y\in \mathcal{Y}$. Then, for $\theta\in \Omega$ and any $y\in \mathcal{Y}$ with $\mathcal{X}\neq \emptyset$, 
$$l(\theta)\geq l(\theta^{(t)}),$$ if, $$Q(\theta\vert \theta^{(t)})\geq Q(\theta^{(t)}\vert \theta^{(t)}).$$
\end{theorem}
\textbf{Theorem \ref{thm:1}} can also be extended to the \textit{maximum a posteriori} (MAP) EM, where we have a prior on $\theta$. This extension is established using the following Theorem. 
\begin{theorem}\label{thm:2}
Let random variables $X$ and $Y$ have parametric densities with parameter $\theta\in \Omega$, where $\theta$ is distributed according to the density $p(\theta)$ on $\Omega$. Suppose the support of $X$ does not depend on $\theta$, and the Markov relationship $\theta\rightarrow X \rightarrow Y$, that is,
\[p(y\vert x,\theta) = p(y\vert x) \]
holds for all $\theta\in \Omega$, $X\in\mathcal{X}$ and $y\in \mathcal{Y}$. Then, for $\theta\in \Omega$ and any $y\in \mathcal{Y}$ with $\mathcal{X}\neq \emptyset$, 
$$l(\theta) + \log p(\theta)\geq l(\theta^{(t)})+ \log p(\theta^{(t)}),$$
if,
$$Q(\theta\vert \theta^{(t)}) + \log p(\theta)\geq Q(\theta^{(t)}\vert \theta^{(t)})+ \log p(\theta^{(t)}).$$
\end{theorem}

Since both the Theorems are taken directly from \cite{gupta2011theory}, we do not include their proofs here, but provide them in Appendix A for the sake of completeness. Now, recall that we assume factorized priors to approximate the posterior. Moreover, consider the graphical model representation of the algorithm in Fig. \ref{fig:GraphicalModel}. We decompose the graphical model into chains that satisfy the assumptions in the two Theorems. These chains are $\mathbf{w}\rightarrow {\mathbf{f}}\rightarrow \mathbf{Y}$, $\mathbf{\alpha}\rightarrow {\mathbf{w}}\rightarrow \mathbf{f}$, $\mathbf{\alpha}\rightarrow {\mathbf{w}}\rightarrow \mathbf{\beta_{\mathbf{f}}}$, and $\mathbf{\beta_{\mathbf{f}}}\rightarrow {\mathbf{f}}\rightarrow \mathbf{Y}$. In some of these chains, we also have priors on the parameters. Therefore, we use \textbf{Theorem \ref{thm:2}} to prove the enhancement of their log-likelihood function by using the EM updates. The corresponding likelihood functions for these chains generate factors in the complete likelihood function (\ref{eq:10}), and all these factors are bounded in their respective domains. We also know that if two functions are monotonic in their domains, the product of these functions is monotonic in its domain. Hence, the overall log-likelihood function is bounded, and its value increases when we apply the updates in each iteration until it converges to the optimal log-likelihood value \cite{gupta2011theory}. 

However, this result does not necessarily guarantee the convergence of the parameters to an optimal value. Indeed, there has not been any general convergence theorem for the EM algorithm. To study the convergence of the parameters, we need more information about the family of the desired distribution and the structure of the log-likelihood function \cite{gunawardana2005convergence}. However, even knowing the family of the desired distribution is not sufficient when we are dealing with the variational EM algorithm with a factorized posterior (as discussed in the Introduction) since there is no guarantee that the estimated parameters correspond to the stationary points of the likelihood function.
\section{Experimental Results} 
\label{sec: Results}
In this section, we present the implementation details and characterize the performance of the Batch-HBLR method on three different stochastic dynamical systems. First, we evaluate our method on a stochastic mass-spring-damper (MSD) system, and then on a more complex double inverted pendulum on a cart (DIPC) system. Last, we validate its effectiveness on the synthetic version of a real-world micro-robotic system.
\subsection{Implementation} \label{sec:Implementation}
In order to construct non-informative priors, we choose $a_0^\alpha= b_0^\alpha= a_0^\beta= b_0^\beta= 10^{-6}$ and $\beta_y = 10^{9}$ as the hyperparameters. We assume similar initial precision ($\boldsymbol{\alpha}_m$) for all the models weights, and set the initial value equal to $\frac{a_0^{\alpha}}{b_0^{\alpha}}\mathds{1}_p$ since they are generated by a gamma distribution. The matrix inversion instability avoidance parameter, $\epsilon$, is chosen to be $10^{-10}$. We use $w_{gen}=0.5$ as the RBF activation threshold. If the RBF value for a sample ($\mathbf{x}^{(n)}$) is less than $w_{gen}$, the algorithm adds a new local model with that sample as the center, and initializes its parameters in the I\small{NITIALIZE}\normalsize LM function in \textbf{Algorithm \ref{alg:main}}. These parameters comprise the length scale ($\lambda_{init} =0.3$), the mean $\mu_{\mathbf{w}_m}=\mathbf{0}$, and the covariance $\boldsymbol{\Sigma}_{\mathbf{w}_m} = \mathbf{0}_{p \times p}$ of the models weights. We only retain those features with precision values less than $1000$. If the precision becomes larger than this threshold, we set the corresponding weight to zero and prune the corresponding feature. The learning rate, $\kappa$, of $\lambda$ is selected to be $0.0001$ and the iteration threshold, $\delta$, is also chosen as 0.0001.

In the two illustrative examples, we report the normalized mean squared error ($nMSE$), which is computed in line 19 of \textbf{Algorithm \ref{alg:testing}}. When we have a response variable ($y$) that is a constant function, $\Var(y)$ is zero, and unless we add noise to the data, we are not able to use $nMSE$ and  compute $MSE$ instead. For the micro-robotic system, some of the states remain constant. Therefore, we use $MSE$ both as a termination criterion in the training algorithm and as a performance measure in the test algorithm. We allow a maximum iteration limit of 200 for training the models. However, the algorithm stops if the difference between the previous and current $nMSE$ values is less than $0.0001$. It is worth mentioning the models are learned before 200 iterations in all the examples. 

Algorithms \ref{alg:main} to \ref{alg:testing} are written in \textbf{Python 2.7} and tested on both \textbf{Windows 10} and \textbf{Ubuntu 16.04 LST} operating systems. We randomly select the training ($67\%$ of the data) and test ($33\%$ of the data) sets using the \textbf{train-test-split} function from the \textbf{scikit-learn} library in \textbf{Python}. The simulator for generating the micro-robot trajectories in Section \ref{Sec: Micro-Robotic System} is written in \textbf{C++} programming language, and is run on the \textbf{Ubuntu 16.04 LST} operating system. The source code and the examples would be made available in \textbf{GitHub} later on. 
\subsection{Illustrative Examples}\label{Illustration} 
\subsubsection{Stochastic Mass and Spring Damper system}
\begin{figure}[H]\centering 
  \includegraphics[width=0.5\columnwidth]{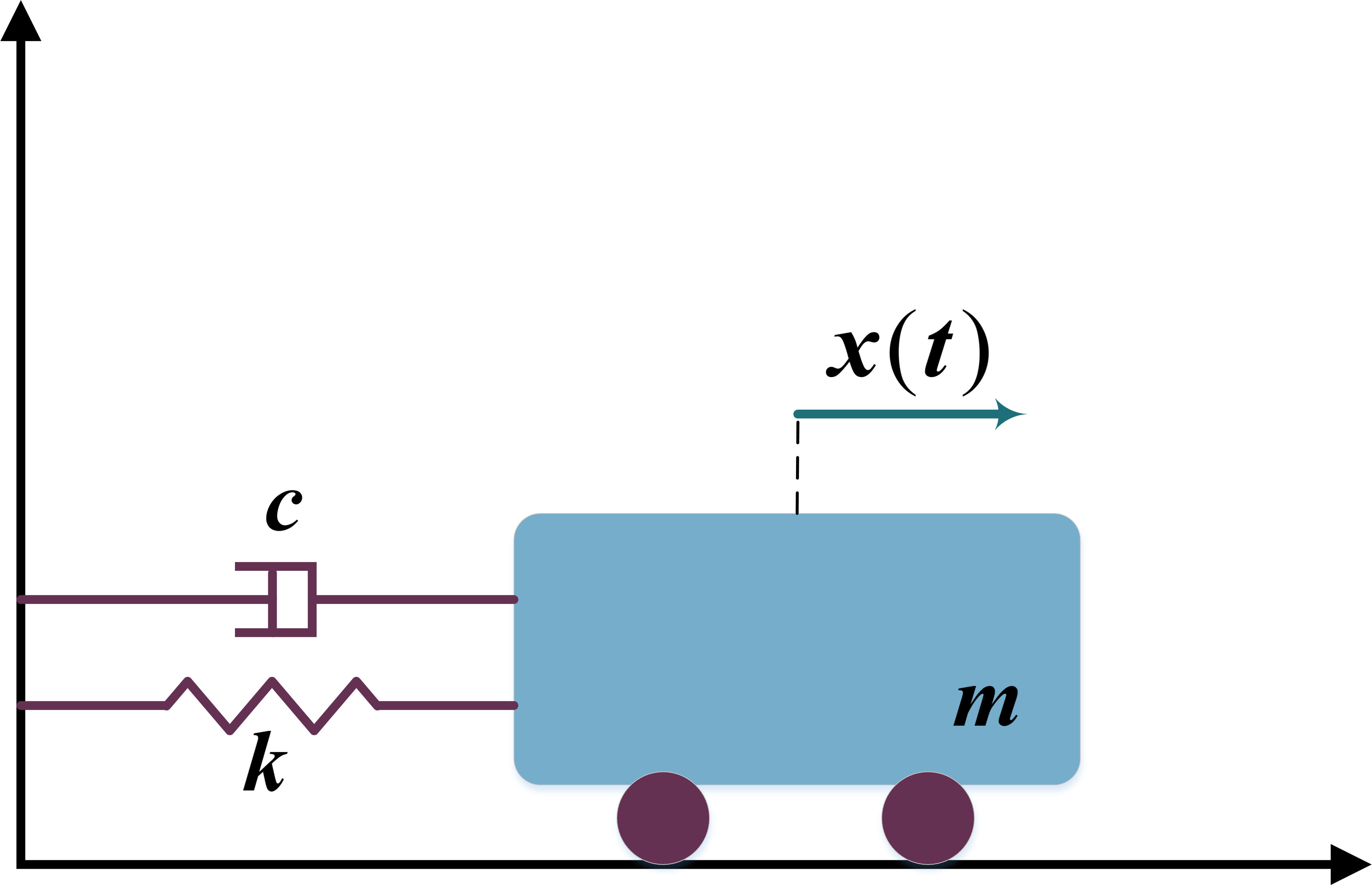}
\caption{Stochastic Mass and Spring Damper system. The damping ratio is $\gamma =1$, and the natural frequency is $\nu =3$ for this system.}
\label{fig:MSD_SYS}       
\end{figure}

\begin{figure}[H]\centering 
  \includegraphics[width=0.75\columnwidth]{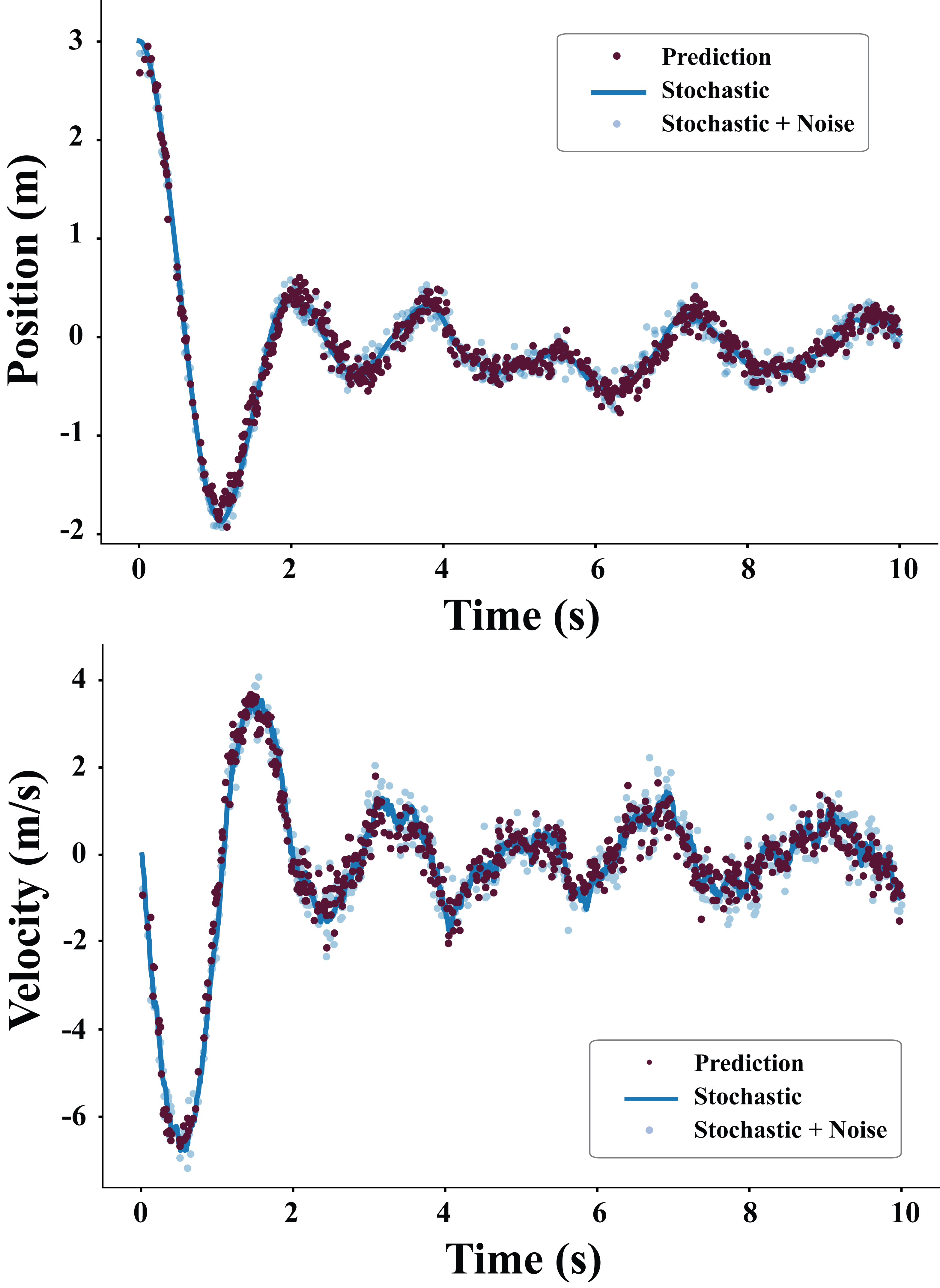}
\caption{Prediction results of our Batch-HBLR method on a stochastic mass-spring damper (MSD) system with varying amounts of noise.} 
\label{fig:MSD}
\end{figure}

We build a stochastic mass-spring damper system by adding white noise to the input of a standard mass-spring damper system (see Fig. \ref{fig:MSD_SYS}). Its linearized state equation is then given by
\begin{equation}\label{eq:50}
\left(\begin{matrix}
\dot{x}_1(t)\\ \dot{x}_2(t)
\end{matrix}\right) = \left(\begin{matrix}
0& 1\\ -\nu^2 & -\gamma
\end{matrix}\right) \left(\begin{matrix}
x_1(t)\\ x_2(t)
\end{matrix}\right) + \left(\begin{matrix}
0\\ 1
\end{matrix}\right) w(t) .
\end{equation}
Equation (\ref{eq:50}) is an It\^o equation \cite{Karatzas2012}, where $x_1$ and $x_2$ are the position and velocity of the mass, respectively, (`` $\dot{}$ '' represents time derivative), and $w\sim \mathcal{N}(0,1)$. In Fig. \ref{fig:MSD_SYS}, $x(t)$ is equivalent to $x_1$ in the state equation, and $k$ and $c$ are the stiffness of the spring and the damping coefficient, respectively. In (\ref{eq:50}), $\nu = \sqrt[]{\frac{k}{m}}$ is the natural frequency of the system, and $\gamma = \frac{c}{m}$ is the damping ratio. It is solved using \textbf{itoint} a numerical integration method in the \textbf{sdeint} Python library. 
\normalsize
The system starts from $\mathbf{x}=\left[\begin{matrix} 
3.0, & 0.0\end{matrix}\right]^\top$ and is simulated for $10$ seconds in increments of $0.005$ second. The effect of measurement noise is included by adding a white noise in the form of $\mathbf{x}(t) = \mathbf{x}(t) + s_d*w$ where, $w\sim \mathcal{N}(0,1)$. $\textit{std}=0.1$ for the first state and $\textit{std}=0.4$ for the second one. The training set for the Batch-HBLR method has size $N=1340$, 3-dimensional input ($\mathbf{X} = [\mathbf{x}[0:N-1]^\top, \mathbf{Time}]$), and 2-dimension response ($\mathbf{Y} = \mathbf{x}[1:N]$). We train the  model on every dimension of the response separately. The predicted results for the test set ($N=660$) using the trained model are shown in Fig. \ref{fig:MSD}. They closely match the ``true'' state values regardless of the amount of noise added into the stochastic system. 

The important quantitative performance measures of the training and test algorithms are summarized in Table \ref{tab:MSD}. The reported prediction time is the time taken by the algorithm to make prediction for one new sample\footnote{Only the average time is reported as the standard error is negligible}. We observe that not only is the algorithm fast in making predictions, it also converges much before reaching the maximum iteration limit during training. The number of local models ($\#$ of LMs) depends only on the inputs and not on the responses ($\mathbf{Y}$). Therefore, it is the same for both the response states. This number is more than two orders of magnitude smaller than the number of samples, making the model extremely parsimonious. The training and test errors are comparable, and very small. The actual expressions of the learned local models for both $x(t)$ and $\dot{x}(t)$ are reported in Tables \ref{tab:MSD_models1} and \ref{tab:MSD_models2} in Appendix B. 

\begin{longtable}{c c c c c c}
\caption{Performance measures of our Batch-HBLR method on the simulated data of a noisy MSD system.}\\
\label{tab:MSD}
 \begin{tabular}{||>{\centering\arraybackslash}p{0.8cm} >{\centering\arraybackslash}p{2cm} >{\centering\arraybackslash}p{2.5cm} >{\centering\arraybackslash}p{1cm} >{\centering\arraybackslash}p{1cm} >{\centering\arraybackslash}p{2cm}||} 
 \hline 
 \multicolumn{6}{|c|}{\cellcolor{Gray} 
 \begin{tabular}{c}\# of training samples: $1340$ \& \# of testing samples: $660$\\Windows 10 - 64 bit  \end{tabular}}\\
 \hline\hline
 States &\centering\begin{tabular}{c}
 nMSE \\(training)
 \end{tabular} & \begin{tabular}{c}
 nMSE \\(test)
 \end{tabular} & \begin{tabular}{c}
 \# of \\LMs
 \end{tabular}  & \begin{tabular}{c}
 \# of \\ iter.
 \end{tabular}& \begin{tabular}{c}
 Prediction \\time (ms)
 \end{tabular} \\ [0.5ex] 
 \hline\hline
  $x(t)$ & 0.04300 & 0.04759 & 3 &  57 & 9\\ 
 \hline
 $\dot{x}(t)$  & 0.08262 & 0.0.08189 & 3  & 30 & 9\\
 \hline
\end{tabular}\\
\end{longtable}

\subsubsection{Stochastic double inverted pendulum on a cart}
\begin{figure}[H]\centering 
  \includegraphics[width=0.7\columnwidth]{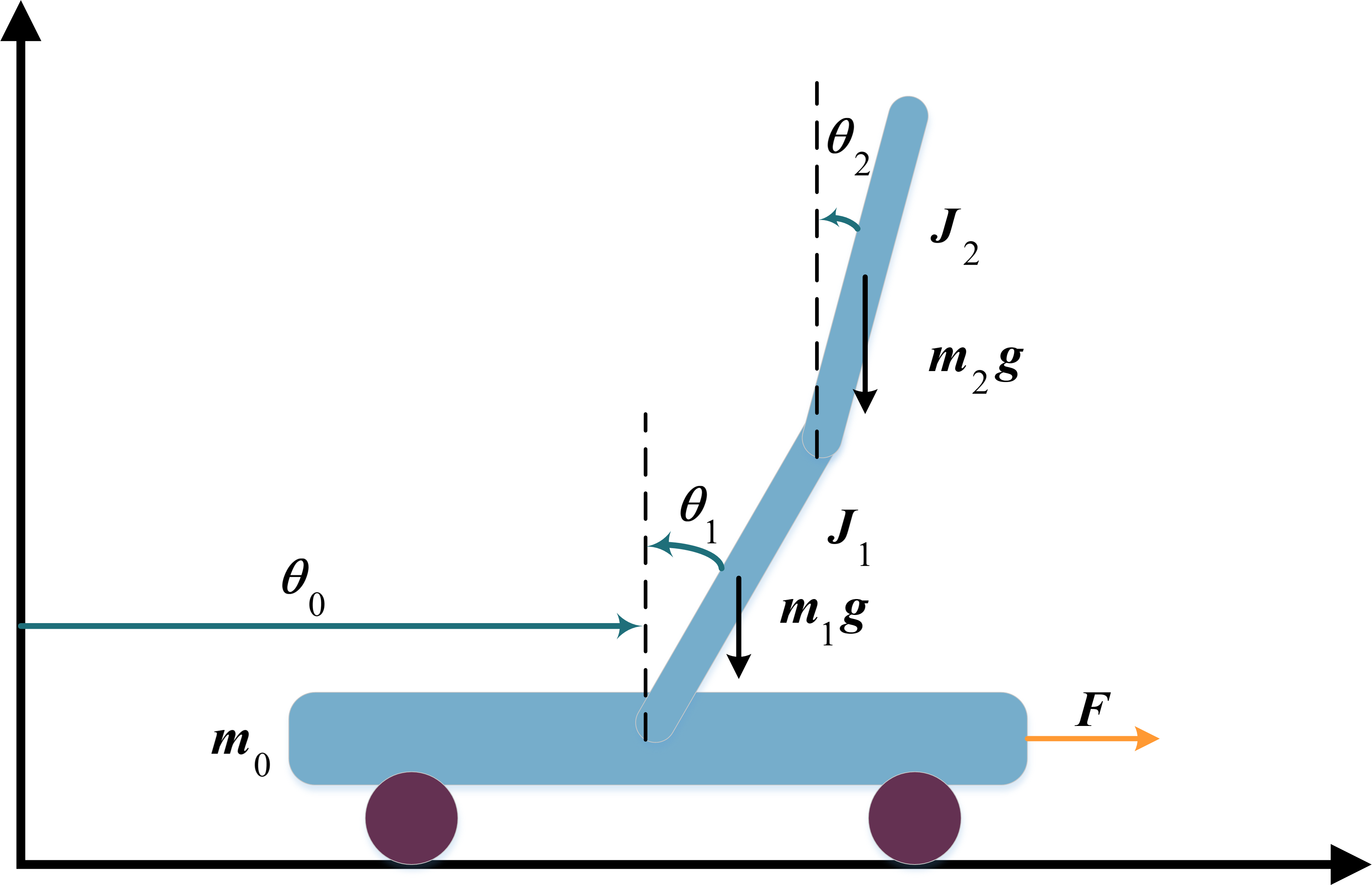}
\caption{Stochastic double inverted pendulum on a cart with masses, $m_0 = 1.5$, $m_1 = 0.5$, and $m_2 = 0.75$. The length of the first pendulum is $l_1=0.5\ \textnormal{m}$ and that of the second pendulum is $l_2=0.75\ \textnormal{m}$. The moment of inertia $J_i=\frac{1}{3}m_i l_i^2$ for $i=1,2$. $\mathbf{F}$ is the input force applied to the cart.}
\label{fig:DIPonC_SYS}      
\end{figure}

Our second example is a stochastic double inverted pendulum on a cart, which is depicted in Fig. \ref{fig:DIPonC_SYS}. The differential equations of the system dynamics is adopted from the model presented in \cite{bogdanov2004optimal}. 
We now provide a brief description of the dynamics model. The state vector is defined as $\mathbf{x} =\left[\begin{matrix} 
\theta_0, \theta_1, \theta_2, \dot{\theta}_0,\dot{\theta}_1,\dot{\theta}_2\end{matrix}\right]^\top$, where the position of the cart is in meters and the angles are in radians. For the system in Fig. \ref{fig:DIPonC_SYS} the state-space equation is
\begin{align}
\dot{\mathbf{x}}(t) &= \mathbf{A}\mathbf{x}(t) + \mathbf{B}\mathbf{F}(t) + \mathbf{D} w(t)\label{51}\\
\mathbf{y}(t) &= \mathbf{x}(t)\label{52}
\end{align}
where,
\begin{align*}
\mathbf{A} &=\left(\begin{matrix}
0.0\quad & 0.0\quad & 0.0\quad & 1.0\quad & 0.0\quad & 0.0\quad\\
0.0\quad & 0.0\quad & 0.0\quad & 0.0\quad & 1.0\quad & 0.0\quad\\
0.0\quad & 0.0\quad & 0.0\quad & 0.0\quad & 0.0\quad & 1.0\quad\\
0.0\quad & 0.0\quad & -7.49\ \ \ & 0.798\ \ & 0.0\quad & 0.0\quad\\
0.0\quad & 0.0\quad & 74.93\ \ & -33.71\ \ \ & 0.0\quad & 0.0\quad\\
0.0\quad & 0.0\quad & -59.94\ \ \ & 52.12\ \ & 0.0\quad & 0.0\quad\\
\end{matrix}\right)\\
\mathbf{B} &= \left(\begin{matrix} 0.0\\ 0.0\\ 0.0\\ -0.61\\ 1.5\\ -0.3
\end{matrix}\right)\\
\mathbf{D} &= \left(\begin{matrix}
0.0\\0.0\\0.0\\ 0.1\\ 0.1\\ 0.1
\end{matrix}\right),
\end{align*}
and $w\sim \mathcal{N}(0,1)$. The stochastic tracking error dynamics is written as
\begin{equation}\label{53}
\dot{\mathbf{e}}(t) = (\mathbf{A} + \mathbf{B}\mathbf{F}(t))\mathbf{e}(t) + \mathbf{D} w(t),
\end{equation}
where, $\mathbf{e}(t) = \mathbf{x}(t) -\mathbf{x}_d(t)$ is the error calculated with respect to the desired trajectory $\mathbf{x}_d(t)$.

\begin{figure}[H]\centering 
  \includegraphics[width=0.9\columnwidth]{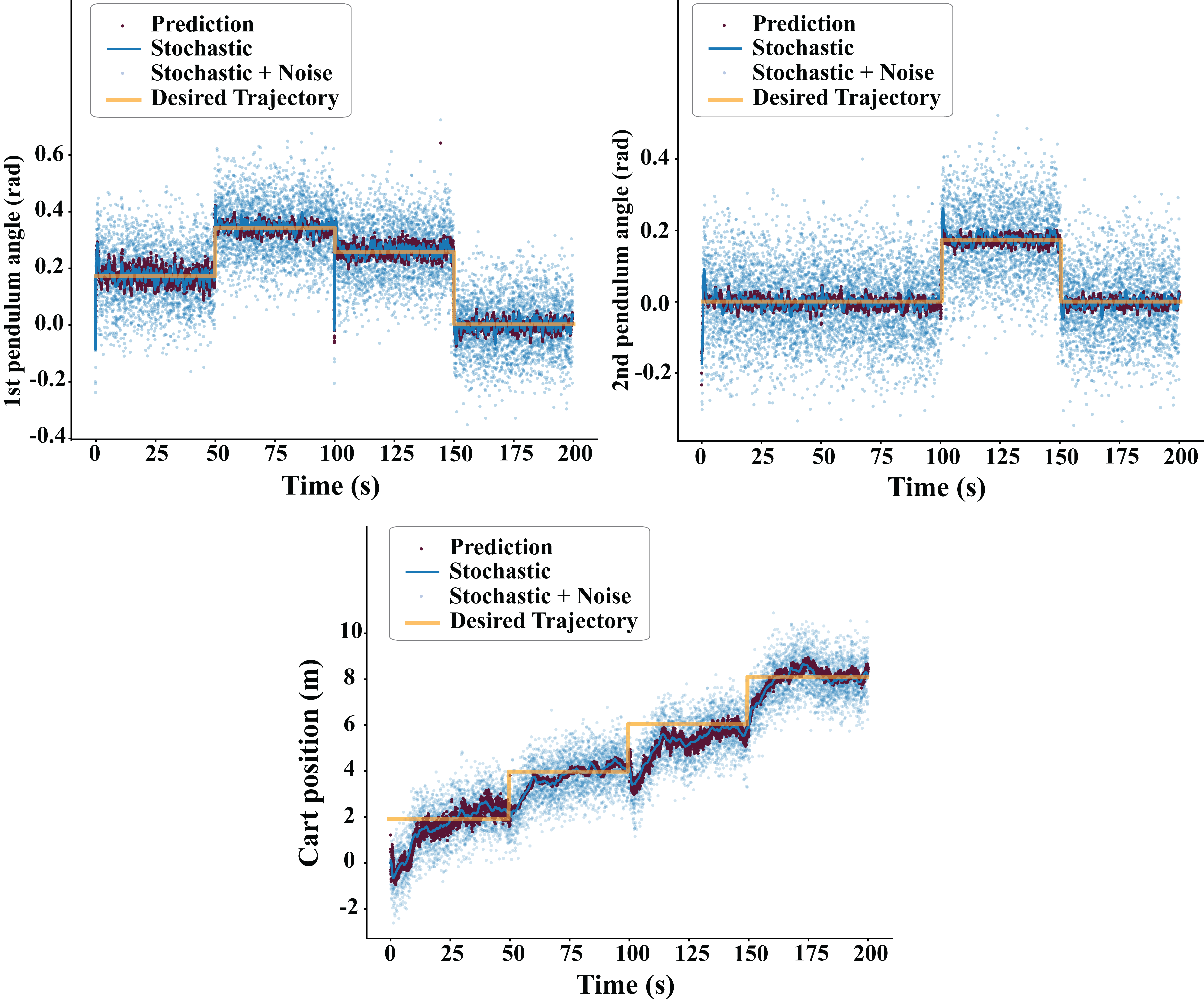}
\caption{Prediction results of our Batch-HBLR method on a stochastic double inverted pendulum on a cart (SDIP) system.} 
\label{fig:SDIP}       
\end{figure}

We consider a Linear Quadratic Regulator (LQR) controller, $\mathbf{F}(t)=-\mathbf{K}\mathbf{x}$, and find it by minimizing a quadratic cost $J$ in the form of $\mathbf{x}^\top\mathbf{Q}\mathbf{x} + \mathbf{F}^\top\mathbf{R}\mathbf{F}$. Here, $\mathbf{K}$ is the solution of the algebraic Riccati equation. 
Defining 
$$\mathbf{Q}=\textbf{diag}\left(\begin{matrix} 
10, & 100, & 100, & 700, & 700, & 700
\end{matrix}\right),$$
and $\mathbf{R}=1$, we get $$\mathbf{K} = \left[\begin{matrix} 
-3.162, &  589.127 , & -842.986, &  -29.493, & 4.469, & -133.079
\end{matrix}\right].$$
The system starts from $\mathbf{x}=\left[\begin{matrix} 
0.0, & 0.175, & -0.175, & 0.0, & 0.0, & 0.0
\end{matrix}\right]^\top$ and follows a desired trajectory for $200$ seconds in increments of $0.01$ second. The effect of measurement noise is incorporated by adding a white noise in the form of $\mathbf{x}(t) = \mathbf{x}(t) + \mathbf{s}_dw$. $\mathbf{s}_d= \left[\begin{matrix} 
0.5,& 0.3,& 0.25,& 0.8,& 0.3,& 0.2\end{matrix}\right]^\top$ is a vector of the standard deviations of the noise we add to the states, and $w\sim \mathcal{N}(0,1)$. The training set for the Batch-HBLR method has $N=13400$ paired samples. The samples have 8-dimensional input space ($\mathbf{X} = [\mathbf{x}[0:N-1]^\top, \mathbf{F}, \mathbf{Time}]$, which consists of the 6 states, the input $\mathbf{F}$, and the simulation time $\mathbf{Time}$. The response state is a 6-dimensional vector $\mathbf{Y} = [\mathbf{x}[1:N]]$, which captures all the states of the system. We train the model on every dimension of the target separately, and report the results in Table \ref{tab:SDIP}. The simulated trajectories of the pendulum and cart used for testing the model are shown in Fig. \ref{fig:SDIP}. The tabulated performance measures and the illustrated prediction results are very similar to those for the stochastic MSD system, indicating the general applicability of our method. The learned local models for one of the response variables $\theta_0(t)$ are reported in Table \ref{tab:SDIP_models} in Appendix B.
\begin{longtable}{c c c c c c}
\caption{Performance measures of our Batch-HBLR method on the simulated data of a noisy SDIP system.}\\
\label{tab:SDIP}
 \begin{tabular}{||>{\centering\arraybackslash}p{0.8cm} >{\centering\arraybackslash}p{2cm} >{\centering\arraybackslash}p{2.5cm} >{\centering\arraybackslash}p{1cm} >{\centering\arraybackslash}p{1cm} >{\centering\arraybackslash}p{2cm}||} 
 \hline 
 \multicolumn{6}{|c|}{\cellcolor{Gray} 
 \begin{tabular}{c}\# of training samples: $13400$ \& \# of testing samples: $6600$\\Windows 10 - 64 bit  \end{tabular}}\\
 \hline\hline
 States &\centering\begin{tabular}{c}
 nMSE \\(training)
 \end{tabular} & \begin{tabular}{c}
 nMSE \\(test)
 \end{tabular} & \begin{tabular}{c}
 \# of \\LMs
 \end{tabular}  & \begin{tabular}{c}
 \# of \\ iter.
 \end{tabular}& \begin{tabular}{c}
 Prediction \\time (ms)
 \end{tabular} \\ [0.5ex] 
 \hline\hline
  $\theta_0$ & 0.0141 & 0.0146 &  3 & 80 &3\\ 
 \hline
 $\theta_1$ & 0.5828 & 0.5821  & 3 & 99 &3\\
 \hline
 $\theta_2$ & 0.9827 &  0.6619 & 3 & 9 &3\\
 \hline
 $\dot{\theta}_0$ & 0.6479 & 0.5687 & 3 & 13 &3\\
 \hline
 $\dot{\theta}_1$ & 0.5718 &  0.5643 & 3 & 33 &3\\
 \hline
 $\dot{\theta}_2$ & 0.9847 & 0.9880 & 3 & 8 &3\\
 \hline
\end{tabular}\\
\end{longtable}
\subsection{Micro-Robotic System}\label{Sec: Micro-Robotic System}
The final and most important example of the paper is a micro-robotic system, in which a microscopic object (robot) is manipulated (actively controlled) in a fluid medium using the optical force produced by shining a tightly focused laser beam at the object. The dynamics of this system is stochastic due to the Langevin force that gives rise to Brownian motion-based diffusion. Rajasekaran \textit{et al.} \cite{Rajasekaran2016} describe the details of the dynamics model, which also includes optical forces, viscous drag, and buoyancy, using a tensor stochastic differential equation. We directly use the OTGO toolbox \cite{OTGO} to generate look-up tables of the optical trapping forces $\mathbf{F}_o$. The dynamics model for one optical trap (local controller generating the optical forces) affecting $n$ spherical micro-robots and other freely diffusing objects, termed as obstacles, is then represented as
\begin{equation}\label{eq:54}
\mathbf{M}\ddot{\mathbf{x}}(t)=\mathbf{F}_o-\mathbf{B}_{d}\dot{\mathbf{x}}(t)-\mathbf{B}_{o}+\mathbf{F}\bf{\eta}.
\end{equation}
The symbols in (\ref{eq:54}) are defined as follows.
\begin{itemize}
    \item $\mathbf{M}$ is the $3n \times 3n$ diagonal mass matrix of the micro-objects.
    \item $\mathbf x$ is a $3n \times 1$ combined vector of the object locations (center coordinates).
    \item $\mathbf{B}_{d}$ is the viscous drag coefficient matrix of dimension\\ $3n \times 3n$ populated by
     $
    \begin{cases}
        6\pi r \mu \quad $ if the object is not close to the cover slide$\\
        \frac{ 6\pi r \mu}{1-\frac{9r}{16h}+\frac{r^3}{8h^3}-\frac{45r^4}{256h^4}-\frac{r^5}{16h^5}}$ if the object is near the cover slide$.\\
    \end{cases}
    $\\
    $\mu$ is the viscosity of the fluid medium and $h$ is the distance between the cover-slip and the object center.
    \item $\mathbf B_o$ is the buoyancy force; it is a $3n \times 3n$ diagonal matrix with $ V\rho_l g$ as the diagonal element in every third row. 
$V$ is the volume of the displaced fluid, $\rho_l$ is the density of the fluid, and $g$ is the acceleration due to gravity.
    \item $\mathbf{F}$ is the diagonal disturbance coupling matrix with dimension $3n \times 3n$, where each term is $\sqrt{2k_bT\gamma}$ with $k_b$ being the Boltzmann constant and $\gamma$ being the viscous drag coefficient. 
    \item $\eta$ is the standard Gaussian variable.
\end{itemize}

\begin{figure}[H]\centering
  \includegraphics[width=0.8\columnwidth]{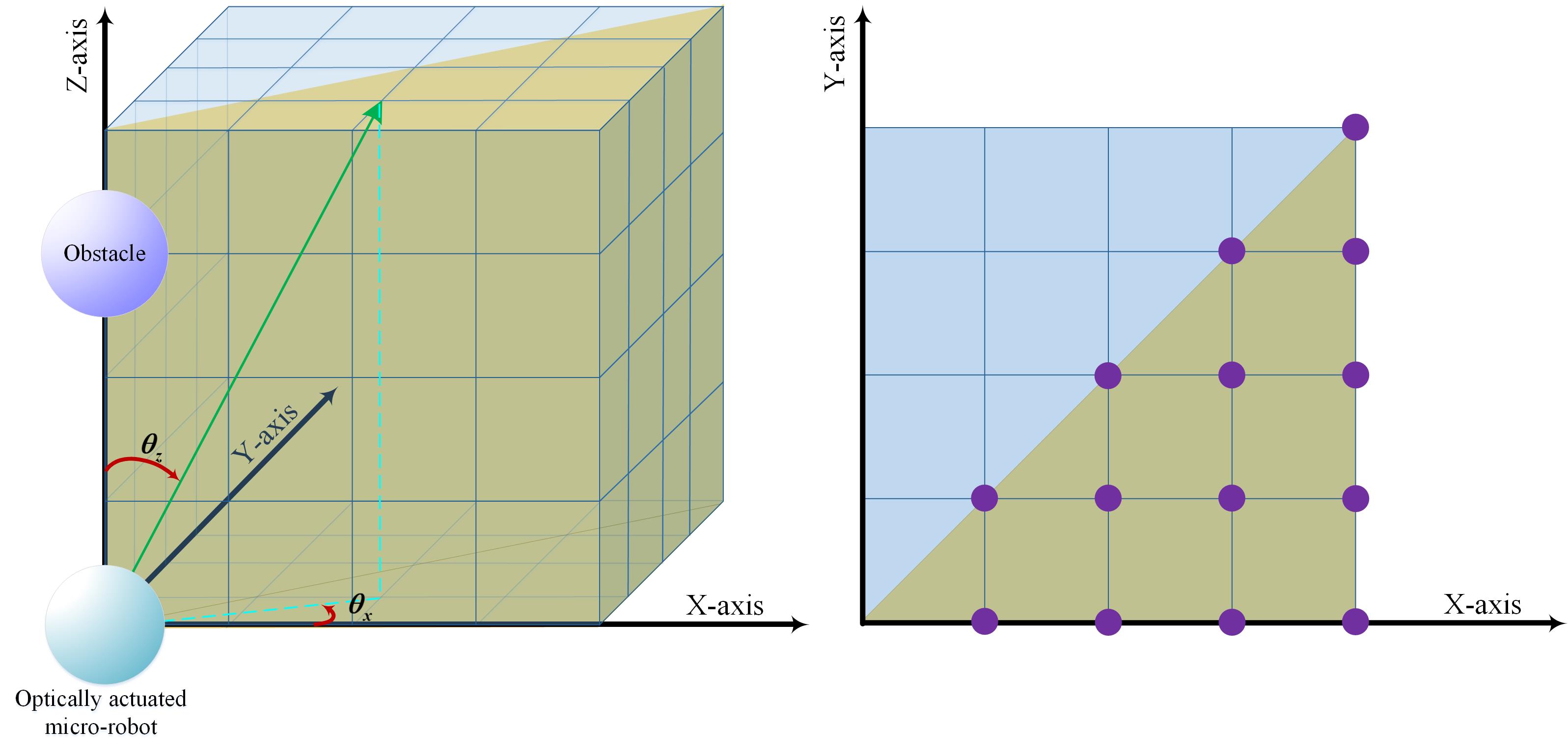}
\caption{In the left figure, the yellow part of the cube shows the volume used for running the experiments (the remaining volume is ignored due to force symmetry considerations). The green arrow is a sample trajectory followed by the controlled micro-robot. $\theta_x$ and $\theta_z$ show the angle made by the trajectory in the X-Y plane with the X-axis and the angle with the Z-axis in the plane defined by the Z-axis and the trajectory vector, respectively. The obstacle is located at the nodes of the 3D grid (see the right figure for the locations on the X-Y plane).} 
\label{fig:FrameDefs}       
\end{figure}

We develop a high-fidelity simulator, akin to the one presented in \cite{banerjee2018step}, to generate the trajectories for various instances of a basic scenario involving exactly one micro-robot and one obstacle in its neighborhood. Note here that this basic scenario, under varying robot velocities and relative displacements of the robot-obstacle pair, is replicable for more complex environments with a larger number of robots and obstacles. We consider a small 3D volume as our environment, which is discretized into a regular $4\times 4\times 4$ grid with its size equal to the robot diameter (Fig. \ref{fig:FrameDefs}). Both the robot and the obstacle consist of a $2$ $\mu$m diameter amorphous silica bead. In all the experiments, the robot starts from the bottom corner of the environment (the origin) and is moved along a straight line with constant speed until it leaves the environment's boundary. The maximum speed at which one can manipulate this robot in a stable manner using a 0.5W laser beam in an aqueous medium at room temperature is 1 $\mu$m/s. We then consider the following discrete set of speeds (in $\mu$m/s) $S=\left\lbrace 0.2, 0.4, 0.6, 0.8, 1.0 \right\rbrace$ 
The optical effect on a micro-robot is symmetric (direction insensitive) in the horizontal plane (X-Y). Therefore, we only simulate the movement in half of the X-Y plane (the shaded region in Fig. \ref{fig:FrameDefs}). The trajectories are generated such that they cover the specified regions within the environment. The angle between the trajectory and the Z-axis is called $\theta_z$ and the angle between the projection of the trajectory in the X-Y plane and the X-axis is termed as $\theta_x$. Five representative directions are used for training and testing the model. Defining the direction of a trajectory with the tuple of $(\theta_x,\theta_z)$, the trajectories $(.,0)$, $(0,45)$, $(45,45)$, $(0,90)$, and $(30,75)$ are used in this experiment. Note that when $\theta_z=0$, the micro-robot is moving along the $Z$ axis; thus, $\theta_x$ is undefined, and we show its direction as $(.,0)$. For every direction of the movement of the micro-robot, we put an obstacle in the environment as shown in Fig. \ref{fig:FrameDefs}. Considering the 3D nature of the environment, we have $14 \times 5 = 70$ obstacle positions. Therefore, we have $70 \times 5 = 350$ simulations for every direction (5 is the size of speed set). 

\begin{figure}[H]\centering 
  \includegraphics[width=0.9\columnwidth]{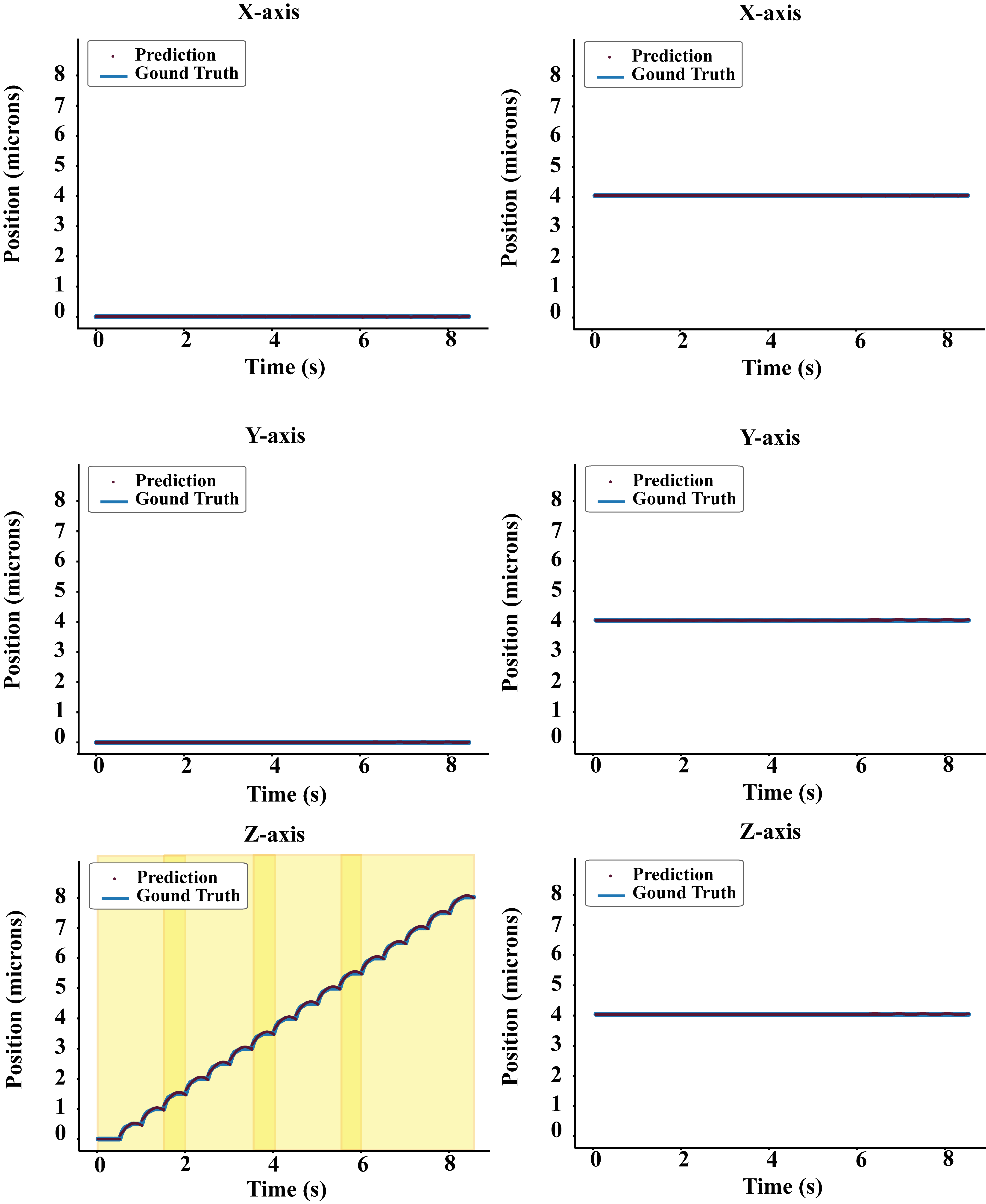}
\caption{Ground truth (simulated) and predicted $X$, $Y$, $Z$ trajectories of a micro-robot (left column) starting from the origin and moving along the $Z$ axis with a speed of 1 $\mu$ m/s. The obstacle (right column) is initially located at the $[4,4,4]$ grid node, and remains unaffected by the moving robot. The bottom left figure also shows the moving window for selecting the four training data batches.}
\label{fig:222}       
\end{figure}

\begin{figure}[H]\centering 
  \includegraphics[width=0.9\columnwidth]{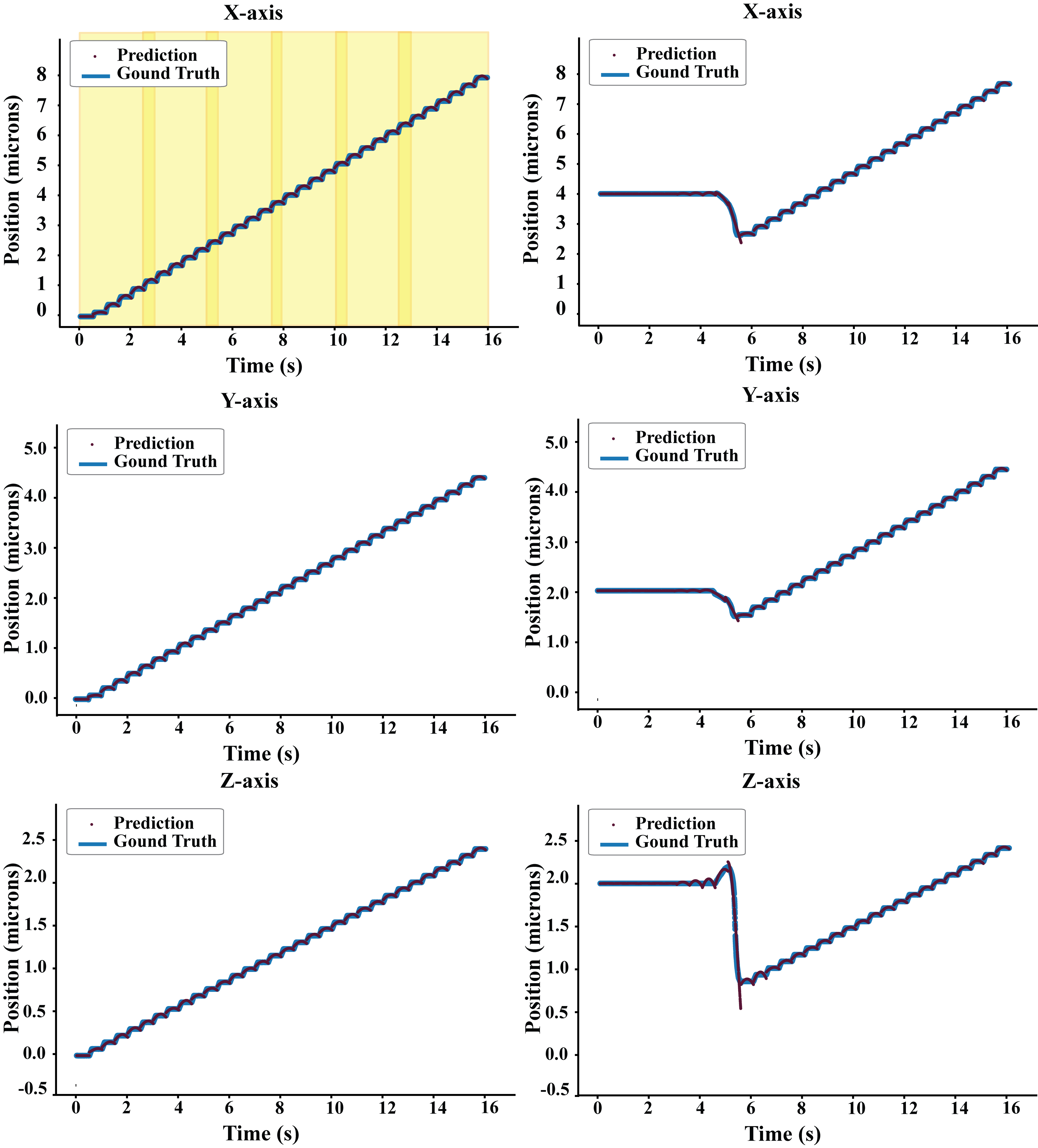}
\caption{Ground truth and predicted $X$, $Y$, $Z$ trajectories of a micro-robot (left column) starting from the origin and moving along the $(30,75)$ direction with a speed of 0.6 $\mu$ m/s. The obstacle is initially located at the $[4,2,2]$ grid node, and gets dragged by the moving robot as it passes close by (right column). The top left figure also shows the moving window for selecting the six training data batches.}
\label{fig:212}       
\end{figure}

Figures \ref{fig:222} and \ref{fig:212} show the test results for two experiments. 
In Fig. \ref{fig:222}, the micro-robot is moving with the maximum speed ($1 \mu m/s$) along the Z-axis and the obstacle is at $[4,4,4]$. The obstacle is far enough in this experiment; therefore, it stays at its initial position. The micro-robot moves along the Z-axis until it leaves the cube. In Fig. \ref{fig:212}, the robot is moving with speed ($0.6 \mu m/s$) along $(30^\circ,75^\circ)$, and the obstacle is at $[4,2,2]$. It gets attracted to the trap after $4$ seconds from the beginning of the experiment, and then the two beads travel together until they leave the environment.

The micro-robot is moved by controlling the position of the optical trap. The simulator updates the trap position every $500 ms$ and stores 100 samples for every control action. As a result, the size of the training set is large. Hence, we partition it into smaller sections with the overlap of one control action between the sections. The number of partitions used for each speed in $S=\left\lbrace 0.2, 0.4, 0.6, 0.8, 1.0 \right\rbrace$ is $P=\left\lbrace 20, 10, 6, 5, 4 \right\rbrace$, respectively.

The average quantitative results with standard errors are presented in Table \ref{tab:Tweezers}. As for every direction, we have 350 experiments, the mean squared error (MSE) is reported along with its statistics. The MSE for the training set is first averaged over the segments and its statistics are then computed for the 350 experiments. The statistics for the total number of local models obtained in each segment are reported in the table under the heading of "$\#$ of LMs". The number of iterations ($\#$ of iter.) is computed similarly to the number of local models.

In addition to accurate training and test results (low MSE), what is significant about this learning algorithm is that it only stores very few samples as the centers of the RBF models. For instance, in the first sub-table in Table \ref{tab:Tweezers}, the average size of the training set is 37200 samples, whereas the average number of local models/RBF centers is 60. This significantly reduces the amount of memory needed for recovering the results and making predictions for new data streams. Moreover, if in some problems one needs to keep updating the model as new data arrives, it is easy to transform this algorithm to operate in an incremental fashion. In this manner, we can continuously expand the model by adding new local models and updating the parameters of all the models accordingly.

Last, but not the least, the algorithm results in fast predictions. The last column of Table \ref{tab:Tweezers} corresponds to the time elapsed while the model makes prediction for a new location. The average is less than 0.5 milliseconds for all the states. This is a very important advantage for a learning algorithm, especially if it is used for learning the dynamics of a robotic system (or the transition probabilities). Fast state estimation not only makes robot motion control more precise but also more adept in reacting to environmental changes.

\newpage
\newpage
\newpage
\begin{longtable}{c c c c c c}
\caption{Performance measures of our Batch-HBLR method for the optically actuated micro-robotic system. The reported values are the average of 350 experiments in which the controlled micro-robot is moved along a particular specified direction. As we partition the training data into $S$ segments and train the HBLR model for each segment separately, the reported training $MSE$ values are the averages of the MSEs observed for all the segments.}\\
\label{tab:Tweezers}
 \begin{tabular}{||>{\centering\arraybackslash}p{0.8cm} >{\centering\arraybackslash}p{2cm} >{\centering\arraybackslash}p{2.5cm} >{\centering\arraybackslash}p{1cm} >{\centering\arraybackslash}p{1cm} >{\centering\arraybackslash}p{2cm}||} 
 \hline 
 \multicolumn{6}{|c|}{\cellcolor{Gray} 
 \begin{tabular}{c}$\theta_z=0^\circ$, $\theta_x$ is not defined \\ Avg. \# of training samples: $37200$\\ Avg. \# of test samples: $18322$\\Ubuntu 16.4  \end{tabular}}\\
 \hline\hline
 States &\centering\begin{tabular}{c}
 Avg. MSE \\(training)
 \end{tabular} & \begin{tabular}{c}
 MSE \\(test)
 \end{tabular} & \begin{tabular}{c}
 \# of \\LMs
 \end{tabular}  & \begin{tabular}{c}
 \# of \\ iter.
 \end{tabular}& \begin{tabular}{c}
 Prediction \\time (ms)
 \end{tabular} \\ [0.5ex] 
 \hline\hline
 $X$ & \begin{tabular}{c}
 $9.93*10^{-11}$\\$\pm 2.09*10^{-13}$
 \end{tabular}  & \begin{tabular}{c}$3.01*10^{-14}$\\$\pm 6.17*10^{-15}$  \end{tabular} & \begin{tabular}{c}$ 60$\\$\pm 35$  \end{tabular}  & \begin{tabular}{c}$ 2$\\$\pm 0$  \end{tabular} &\begin{tabular}{c}$ 0.27$\\$\pm 0.07$  \end{tabular} \\ 
 \hline
 $Y$ &  \begin{tabular}{c}$1.00*10^{-10}$\\$\pm 1.00*10^{-12} $\end{tabular} &\begin{tabular}{c} $3.84*10^{-14}$\\$\pm 1.27*10^{-14}$\end{tabular}   & \begin{tabular}{c}$ 60$\\$\pm 35$  \end{tabular} &  \begin{tabular}{c}$ 2$\\$\pm 0$  \end{tabular} & \begin{tabular}{c}$ 0.27$\\$\pm 0.07$  \end{tabular} \\
 \hline
 $Z$ & \begin{tabular}{c}$1.07*10^{-3}$\\$\pm 1.94*10^{-4}$ \end{tabular} & \begin{tabular}{c} $1.06*10^{-3}$\\$\pm 1.81*10^{-4}$  \end{tabular} &  \begin{tabular}{c}$ 60$\\$\pm 35$  \end{tabular}  & \begin{tabular}{c}$ 62$\\$\pm 3$  \end{tabular}& \begin{tabular}{c}$ 0.27$\\$\pm 0.07$  \end{tabular}\\
 \hline
 $X_{obs}$ & \begin{tabular}{c}$1.27*10^{-3}$\\$\pm 1.16*10^{-3}$ \end{tabular} &\begin{tabular}{c} $1.24*10^{-3}$\\$\pm 1.13*10^{-3}$  \end{tabular} & \begin{tabular}{c}$ 60$\\$\pm 35$  \end{tabular}  & \begin{tabular}{c}$ 59$\\$\pm 23$  \end{tabular} & \begin{tabular}{c}$ 0.27$\\$\pm 0.07$  \end{tabular}\\
 \hline
 $Y_{obs}$ & \begin{tabular}{c}$1.10*10^{-3}$\\$\pm 1.32*10^{-3}$ \end{tabular} & \begin{tabular}{c}$1.08*10^{-3}$\\$\pm 1.29*10^{-3}$ \end{tabular} & \begin{tabular}{c}$ 60$\\$\pm 35$  \end{tabular}  & \begin{tabular}{c}$ 41$\\$\pm 28$  \end{tabular} &  \begin{tabular}{c}$ 0.27$\\$\pm 0.07$  \end{tabular}\\
 \hline
 $Z_{obs}$ & \begin{tabular}{c}$1.30*10^{-3}$\\$\pm 1.18*10^{-3}$  \end{tabular} &\begin{tabular}{c} $1.28*10^{-3}$\\$\pm 1.15*10^{-3}$  \end{tabular} &\begin{tabular}{c}$ 60$\\$\pm 35$  \end{tabular} & \begin{tabular}{c}$ 55$\\$\pm 24$  \end{tabular}& \begin{tabular}{c}$ 0.27$\\$\pm 0.07$  \end{tabular}\\[1ex] 
 \hline
\end{tabular}\\ \\
\begin{tabular}{||>{\centering\arraybackslash}p{0.8cm} >{\centering\arraybackslash}p{2cm} >{\centering\arraybackslash}p{2.5cm} >{\centering\arraybackslash}p{1cm} >{\centering\arraybackslash}p{1cm} >{\centering\arraybackslash}p{2cm}||} 
 \hline
 \multicolumn{6}{|c|}{ \cellcolor{Gray}\begin{tabular}{c}$\theta_z=90^\circ$, $\theta_x=0^\circ$ \\ Avg. \# of training samples: $37400$\\ Avg. \# of test samples: $18421$\\Ubuntu 16.4 - 64 bit  \end{tabular}} \\
 \hline\hline
 States &\begin{tabular}{c}
 Avg. MSE \\(training)
 \end{tabular} & \begin{tabular}{c}
 MSE \\(test)
 \end{tabular} & \begin{tabular}{c}
 \# of \\LMs
 \end{tabular}  & \begin{tabular}{c}
 \# of \\ iter.
 \end{tabular}& \begin{tabular}{c}
 Prediction \\time (ms)
 \end{tabular} \\ [0.5ex] 
 \hline\hline
 $X$ &  \begin{tabular}{c}$ 1.90*10^{-3}$\\$\pm 7.76*10^{-4} $  \end{tabular} &  \begin{tabular}{c}$ 1.90*10^{-3}$\\$\pm 7.67*10^{-4} $  \end{tabular} &  \begin{tabular}{c}$58 $\\$\pm 37$\end{tabular}  & \begin{tabular}{c}$60 $\\$\pm 4$  \end{tabular} & \begin{tabular}{c}$0.29 $\\$\pm 0.09$  \end{tabular}\\ 
 \hline
 $Y$ &  \begin{tabular}{c}$9.98*10^{-11} $\\$\pm 9.47*10^{-13}  $  \end{tabular} &  \begin{tabular}{c}$ 6.28*10^{-14}$\\$\pm 3.11*10^{-14} $  \end{tabular}  & \begin{tabular}{c}$58 $\\$\pm 37$ \end{tabular}  & \begin{tabular}{c}$ 2$\\$\pm 0$  \end{tabular} & \begin{tabular}{c}$0.29 $\\$\pm 0.09$  \end{tabular}\\
 \hline
 $Z$ & \begin{tabular}{c}$5.90*10^{-6} $\\$\pm 1.84*10^{-7} $  \end{tabular}  &  \begin{tabular}{c}$6.01*10^{-6} $\\$\pm 3.37*10^{-7} $  \end{tabular}  & \begin{tabular}{c}$58 $\\$\pm 37$ \end{tabular}  & \begin{tabular}{c}$ 2$\\$\pm 0 $  \end{tabular} & \begin{tabular}{c}$0.29 $\\$\pm 0.09$ \end{tabular}\\
 \hline
 $X_{obs}$ & \begin{tabular}{c}$ 3.60*10^{-3}$\\$\pm 2.36*10^{-3} $  \end{tabular}  & \begin{tabular}{c}$3.60*10^{-3} $\\$\pm 2.35*10^{-3} $  \end{tabular}  & \begin{tabular}{c}$58 $\\$\pm 37$ \end{tabular} & \begin{tabular}{c}$75 $\\$\pm 9$  \end{tabular} & \begin{tabular}{c}$0.29 $\\$\pm 0.09$ \end{tabular}\\
 \hline
 $Y_{obs}$ & \begin{tabular}{c}$3.33*10^{-3} $\\$\pm 2.50*10^{-3} $  \end{tabular}  & \begin{tabular}{c}$ 3.32*10^{-3}$\\$\pm 2.49*10^{-3} $  \end{tabular} & \begin{tabular}{c}$58 $\\$\pm 37$\end{tabular} & \begin{tabular}{c}$51 $\\$\pm 28$  \end{tabular} & \begin{tabular}{c}$0.29 $\\$\pm 0.09$ \end{tabular} \\
 \hline
 $Z_{obs}$ & \begin{tabular}{c}$ 3.52*10^{-3}$\\$\pm 2.48*10^{-3}$  \end{tabular}  &  \begin{tabular}{c}$3.51*10^{-3} $\\$\pm 2.47*10^{-3} $  \end{tabular} & \begin{tabular}{c}$58 $\\$\pm 37$\end{tabular} & \begin{tabular}{c}$61$\\$\pm 25$  \end{tabular} & \begin{tabular}{c}$0.29 $\\$\pm 0.09$\end{tabular}\\[1ex] 
 \hline
\end{tabular}\\ \\
\begin{tabular}{||>{\centering\arraybackslash}p{0.8cm} >{\centering\arraybackslash}p{2cm} >{\centering\arraybackslash}p{2.5cm} >{\centering\arraybackslash}p{1cm} >{\centering\arraybackslash}p{1cm} >{\centering\arraybackslash}p{2cm}||} 
 \hline
 \multicolumn{6}{|c|}{ \cellcolor{Gray}  \begin{tabular}{c}$\theta_z=45^\circ$, $\theta_x=0^\circ$ \\ Avg. \# of training samples: $52200$\\ Avg. \# of test samples: $25710$\\Ubuntu 16.4  \end{tabular}} \\
 \hline\hline
 States &\begin{tabular}{c}
 Avg. MSE \\(training)
 \end{tabular} & \begin{tabular}{c}
 MSE \\(test)
 \end{tabular} & \begin{tabular}{c}
 \# of \\LMs
 \end{tabular}  & \begin{tabular}{c}
 \# of \\ iter.
 \end{tabular}& \begin{tabular}{c}
 Prediction \\time (ms)
 \end{tabular} \\ [0.5ex] 
 \hline\hline
 $X$ &  \begin{tabular}{c}$ 1.78*10^{-3}$\\$\pm 4.31*10^{-4} $  \end{tabular} &  \begin{tabular}{c}$ 1.75*10^{-3}$\\$\pm 4.08*10^{-4} $  \end{tabular} &  \begin{tabular}{c}$ 81$\\$\pm 46$  \end{tabular}  & \begin{tabular}{c}$ 85$\\$\pm 5 $  \end{tabular} & \begin{tabular}{c}$ 0.36$\\$\pm 0.07$  \end{tabular}\\ 
 \hline
 $Y$ &  \begin{tabular}{c}$9.97*10^{-11} $\\$\pm 7.495*10^{-13} $  \end{tabular} &  \begin{tabular}{c}$7.40*10^{-14} $\\$\pm 7.67*10^{-14} $  \end{tabular}  & \begin{tabular}{c}$81$\\$\pm 46$  \end{tabular}  & \begin{tabular}{c}$ 2$\\$\pm 0$  \end{tabular} & \begin{tabular}{c}$0.36$\\$\pm 0.07$  \end{tabular}\\
 \hline
 $Z$ & \begin{tabular}{c}$1.23*10^{-3} $\\$\pm 9.41*10^{-5} $  \end{tabular}  &  \begin{tabular}{c}$1.21*10^{-3} $\\$\pm 8.97*10^{-5} $  \end{tabular}  & \begin{tabular}{c}$ 81$\\$\pm 46$  \end{tabular}  & \begin{tabular}{c}$ 85$\\$\pm 5$  \end{tabular} & \begin{tabular}{c}$ 0.36$\\$\pm 0.07$  \end{tabular}\\
 \hline
 $X_{obs}$ & \begin{tabular}{c}$ 3.90*10^{-3}$\\$\pm 2.30*10^{-3} $  \end{tabular}  & \begin{tabular}{c}$3.87*10^{-3} $\\$\pm 2.26*10^{-3} $  \end{tabular}  & \begin{tabular}{c}$ 81$\\$\pm 46$  \end{tabular} & \begin{tabular}{c}$103 $\\$\pm 10 $  \end{tabular} & \begin{tabular}{c}$ 0.36$\\$\pm 0.07$  \end{tabular}\\
 \hline
 $Y_{obs}$ & \begin{tabular}{c}$ 3.54*10^{-3}$\\$\pm 2.60*10^{-3} $  \end{tabular}  & \begin{tabular}{c}$ 3.51*10^{-3}$\\$\pm 2.57*10^{-3} $  \end{tabular} & \begin{tabular}{c}$ 81$\\$\pm 46$  \end{tabular} & \begin{tabular}{c}$65 $\\$\pm 39$  \end{tabular} & \begin{tabular}{c}$ 0.36$\\$\pm 0.07$  \end{tabular} \\
 \hline
 $Z_{obs}$ & \begin{tabular}{c}$ 3.94*10^{-3}$\\$\pm 2.32*10^{-3} $  \end{tabular}  &  \begin{tabular}{c}$ 3.91*10^{-3}$\\$\pm 2.31*10^{-3} $  \end{tabular} & \begin{tabular}{c}$ 81$\\$\pm 46$  \end{tabular} & \begin{tabular}{c}$92 $\\$\pm 22$  \end{tabular} & \begin{tabular}{c}$ 0.36$\\$\pm 0.07$  \end{tabular}\\[1ex] 
 \hline
\end{tabular}\\ \\
\begin{tabular}{||>{\centering\arraybackslash}p{0.8cm} >{\centering\arraybackslash}p{2cm} >{\centering\arraybackslash}p{2.5cm} >{\centering\arraybackslash}p{1cm} >{\centering\arraybackslash}p{1cm} >{\centering\arraybackslash}p{2cm}||} 
 \hline
 \multicolumn{6}{|c|}{ \cellcolor{Gray} \begin{tabular}{c}$\theta_z=45^\circ$, $\theta_x=45^\circ$ \\ Avg. \# of training samples: $74000$\\ Avg. \# of test samples: $36448$\\Windows 10 - 64 bit  \end{tabular}} \\
 \hline\hline
 States &\begin{tabular}{c}
 Avg. MSE \\(training)
 \end{tabular} & \begin{tabular}{c}
 MSE \\(test)
 \end{tabular} & \begin{tabular}{c}
 \# of \\LMs
 \end{tabular}  & \begin{tabular}{c}
 \# of \\ iter.
 \end{tabular}& \begin{tabular}{c}
 Prediction \\time (ms)
 \end{tabular} \\ [0.5ex] 
 \hline\hline
 $X$ &  \begin{tabular}{c}$ 1.99*10^{-3}$\\$\pm 2.24*10^{-4} $  \end{tabular} &  \begin{tabular}{c}$1.99*10^{-3} $\\$\pm 2.31*10^{-4} $  \end{tabular} &  \begin{tabular}{c}$ 111$\\$\pm65 $  \end{tabular}  & \begin{tabular}{c}$108 $\\$\pm 8 $  \end{tabular} & \begin{tabular}{c}$0.47 $\\$\pm 1.96 $\end{tabular}\\ 
 \hline
 $Y$ &  \begin{tabular}{c}$ 2.00*10^{-3}$\\$\pm  2.24*10^{-4}$  \end{tabular} &  \begin{tabular}{c}$ 1.99*10^{-3}$\\$\pm 2.30*10^{-4} $  \end{tabular}  & \begin{tabular}{c}$ 11$\\$\pm 65$  \end{tabular}  & \begin{tabular}{c}$ 108$\\$\pm 8$  \end{tabular} & \begin{tabular}{c}$0.47 $\\$\pm 1.96 $\end{tabular}\\
 \hline
 $Z$ & \begin{tabular}{c}$ 1.63*10^{-3}$\\$\pm 1.19*10^{-4}$  \end{tabular}  &  \begin{tabular}{c}$1.16*10^{-3} $\\$\pm 1.16*10^{-4} $  \end{tabular}  & \begin{tabular}{c}$111 $\\$\pm 65$  \end{tabular}  & \begin{tabular}{c}$ 117$\\$\pm 8$  \end{tabular} & \begin{tabular}{c}$0.47 $\\$\pm 1.96 $\end{tabular}\\
 \hline
 $X_{obs}$ & \begin{tabular}{c}$ 4.15*10^{-3}$\\$\pm 1.90*10^{-3} $  \end{tabular}  & \begin{tabular}{c}$4.19*10^{-3} $\\$\pm 1.93*10^{-3} $  \end{tabular}  & \begin{tabular}{c}$111 $\\$\pm 65$  \end{tabular} & \begin{tabular}{c}$130 $\\$\pm 14$  \end{tabular} & \begin{tabular}{c}$0.47 $\\$\pm 1.96 $ \end{tabular}\\
 \hline
 $Y_{obs}$ & \begin{tabular}{c}$ 4.04*10^{-3}$\\$\pm 1.90*10^{-3} $  \end{tabular}  & \begin{tabular}{c}$ 4.07*10^{-3}$\\$\pm 1.92*10^{-3} $  \end{tabular} & \begin{tabular}{c}$ 111$\\$\pm 65$  \end{tabular} & \begin{tabular}{c}$ 100$\\$\pm 37$  \end{tabular} & \begin{tabular}{c}$0.47 $\\$\pm 1.96 $  \end{tabular} \\
 \hline
 $Z_{obs}$ & \begin{tabular}{c}$ 4.24*10^{-3}$\\$\pm 1.95*10^{-3} $  \end{tabular}  &  \begin{tabular}{c}$ 4.26*10^{-3}$\\$\pm 1.97*10^{-3} $  \end{tabular} & \begin{tabular}{c}$ 111$\\$\pm 65$  \end{tabular} & \begin{tabular}{c}$118 $\\$\pm 26 $  \end{tabular} & \begin{tabular}{c}$0.47 $\\$\pm 1.96 $  \end{tabular}\\[1ex] 
 \hline
\end{tabular}\\ \\
\begin{tabular}{||>{\centering\arraybackslash}p{0.8cm} >{\centering\arraybackslash}p{2cm} >{\centering\arraybackslash}p{2.5cm} >{\centering\arraybackslash}p{1cm} >{\centering\arraybackslash}p{1cm} >{\centering\arraybackslash}p{2cm}||} 
 \hline
 \multicolumn{6}{|c|}{\cellcolor{Gray} \begin{tabular}{c}$\theta_z=75^\circ$, $\theta_x=30^\circ$ \\ Avg. \# of training samples: $44000$\\ Avg. \# of test samples: $21672$\\ Ubuntu 16.4  \end{tabular} } \\
 \hline\hline
 States &\begin{tabular}{c}
 Avg. MSE \\(training)
 \end{tabular} & \begin{tabular}{c}
 MSE \\(test)
 \end{tabular} & \begin{tabular}{c}
 \# of \\LMs
 \end{tabular}  & \begin{tabular}{c}
 \# of \\ iter.
 \end{tabular}& \begin{tabular}{c}
 Prediction \\time (ms)
 \end{tabular} \\ [0.5ex] 
 \hline\hline
 $X$ & \begin{tabular}{c}$1.79*10^{-3} $\\$\pm 7.05*10^{-4} $  \end{tabular}  & \begin{tabular}{c}$1.77*10^{-3} $\\$\pm 6.81*10^{-4} $  \end{tabular}  & \begin{tabular}{c}$ 66$\\$\pm 40$  \end{tabular}   &  \begin{tabular}{c}$ 69$\\$\pm 1$  \end{tabular} & \begin{tabular}{c}$0.32 $\\$\pm 0.06$  \end{tabular} \\ 
 \hline
 $Y$ &  \begin{tabular}{c}$1.30*10^{-3} $\\$\pm  2.15*10^{-4}$  \end{tabular} & \begin{tabular}{c}$1.26*10^{-3}$\\$\pm 1.95*10^{-4} $  \end{tabular}   & \begin{tabular}{c}$66 $\\$\pm 40$  \end{tabular}  & \begin{tabular}{c}$ 61$\\$\pm 2 $  \end{tabular} & \begin{tabular}{c}$0.32 $\\$\pm 0.06$  \end{tabular}\\
 \hline
 $Z$ &  \begin{tabular}{c}$ 1.26*10^{-3}$\\$\pm 5.13*10^{-5} $  \end{tabular} & \begin{tabular}{c}$1.21*10^{-3} $\\$\pm 5.77*10^{-5} $  \end{tabular}   & \begin{tabular}{c}$66 $\\$\pm 40$  \end{tabular}  & \begin{tabular}{c}$50 $\\$\pm 2$  \end{tabular} & \begin{tabular}{c}$ 0.32$\\$\pm 0.06$  \end{tabular}\\
 \hline
 $X_{obs}$ &  \begin{tabular}{c}$ 3.22*10^{-3}$\\$\pm 1.72*10^{-3} $  \end{tabular} &  \begin{tabular}{c}$3.18*10^{-3} $\\$\pm 1.69*10^{-3} $  \end{tabular} & \begin{tabular}{c}$66 $\\$\pm 40$  \end{tabular} & \begin{tabular}{c}$ 86$\\$\pm 8$  \end{tabular} & \begin{tabular}{c}$0.32 $\\$\pm 0.06$  \end{tabular}\\
 \hline
 $Y_{obs}$ & \begin{tabular}{c}$ 3.11*10^{-3}$\\$\pm 1.76*10^{-3} $  \end{tabular}  & \begin{tabular}{c}$ 3.07*10^{-3}$\\$\pm 1.72*10^{-3} $  \end{tabular} & \begin{tabular}{c}$66 $\\$\pm 40$  \end{tabular} & \begin{tabular}{c}$64 $\\$\pm 25$  \end{tabular} & \begin{tabular}{c}$0.32 $\\$\pm 0.06$  \end{tabular} \\
 \hline
 $Z_{obs}$ & \begin{tabular}{c}$ 3.24*10^{-3}$\\$\pm  1.77*10^{-3}$  \end{tabular}  &  \begin{tabular}{c}$3.19*10^{-3} $\\$\pm 1.75*10^{-3} $  \end{tabular} & \begin{tabular}{c}$66 $\\$\pm40 $  \end{tabular} & \begin{tabular}{c}$76 $\\$\pm 18$  \end{tabular} & \begin{tabular}{c}$0.32 $\\$\pm 0.06$  \end{tabular} \\[1ex] 
 \hline
\end{tabular}
\end{longtable}

\section{Conclusions}
\label{Conclusions} 
Learning the state transition dynamics for a robotic system, and being able to make fast and accurate predictions is critical for any real-time motion planning and control problem. The performance of the learning algorithm becomes even more important when stochasticity plays an important role in the system. In this paper, we present a hierarchical Bayesian linear regression model with local features for stochastic dynamics approximation, and demonstrate its performance in three different systems. The model is based on a top-down approach that benefits both from the efficiency of local models and the accuracy of global models.

As the data collected from the interactions of robots with their environment are typically enormous, we need a parsimonious learning model for compact storage. In our case, only a few samples are stored and used as the local RBF feature centers. This is done by optimizing the RBF length scale to obtain the minimum number of local models that best represent the data. Moreover, usually the robots' state vectors are large, and the curse of dimensionality slows down learning and prediction substantially. This emphasizes the importance of a sparse learning algorithm, meaning that the weights of the model are pruned if they do not contribute significantly toward the predictions. The presented algorithm adopts the idea of automatic relevance determination and only retains those weights with significant values. The algorithm is also guaranteed to converge to the optimal values of the log-likelihood function for a factorized representation of the posterior and a Markov relationship among the predictor variables and their parameters. 

These characteristics result in satisfactory performance (low prediction errors, small prediction times, few local models, and consistent convergence in a reasonable number of iterations) on all the three test systems. Hence, we believe that our model would be suitable for optimal control policy generation using a paradigm such as model-based reinforcement learning. Such use would require us to construct the regression models for a denser set of trajectories that would cover the entire operating environments of the robots with additional controlled robots and obstacles. The challenges then would be to train all the models in a reasonable amount of time (with acceptable sample complexity) and identify the correct prediction model efficiently at run time.
\section{Appendix A}
\textbf{Theorem \ref{thm:1}} is proved as follows.
\begin{proof}
Consider the log-likelihood function $l(\theta)=\log p(y\vert \theta)$:
\begin{align*}
l(\theta)&=\log p(y\vert \theta)\\
&= \log \int_{\mathcal{X}(y)}p(x,y\vert \theta) dx\\
&= \log\int_{\mathcal{X}(y)}\frac{p(x,y\vert \theta)}{p(x\vert y, \theta^{(t)})}p(x\vert y, \theta^{(t)}) dx \numberthis \label{eq:50}\\
&= \log \mathbb{E}_{{X}\vert y, \theta^{(t)}}\left[\frac{p(X,y\vert \theta)}{p(X\vert y, \theta^{(t)})}\right]\\
& \textrm{(rewrite the integral as an expectation)}\\
&\geq \mathbb{E}_{{X}\vert y, \theta^{(t)}}\left[\log \frac{p(X,y\vert \theta)}{p(X\vert y, \theta^{(t)})}\right]\\
& \textrm{(by Jensen's inequality)}\\
&= \mathbb{E}_{{X}\vert y, \theta^{(t)}}\left[\log \frac{p(X\vert \theta)p(y\vert X)}{p(X\vert \theta^{(t)})p(y\vert X)\/p(y\vert \theta^{(t)})}\right]\\
& \textrm{(by Bayes' rule and Markov property)}\\
&= \mathbb{E}_{{X}\vert y, \theta^{(t)}}\left[\log \frac{p(X\vert \theta)p(y\vert \theta^{(t)})}{p(X\vert \theta^{(t)})}\right]\\
&= \mathbb{E}_{{X}\vert y, \theta^{(t)}}\left[p(X\vert \theta)\right]-\mathbb{E}_{{X}\vert y, \theta^{(t)}}\left[p(X\vert \theta^{(t)})\right]+\log p(y\vert \theta^{(t)}) \\
&= Q(\theta\vert\theta^{(t)}) - Q(\theta^{(t)}\vert\theta^{(t)}) + l(\theta^{(t)}),\numberthis \label{eq:51}
\end{align*}
where the $Q$-function is defined in (\ref{eq:24}). The assumptions of the base $X$ being independent of $\theta$ and Markov property are needed for the proof. The lower bound of the log-likelihood, therefore, is,
\begin{equation}\label{eq:52}
l(\theta) \geq l(\theta^{(t)}) + Q(\theta\vert\theta^{(t)}) - Q(\theta^{(t)}\vert\theta^{(t)}).
\end{equation}
Based on the assumption $Q(\theta\vert \theta^{(t)})\geq Q(\theta^{(t)}\vert \theta^{(t)})$, we conclude that,
\begin{equation}\label{eq:53}
l(\theta) \geq l(\theta^{(t)}) + Q(\theta\vert\theta^{(t)}) - Q(\theta^{(t)}\vert\theta^{(t)})\geq l(\theta^{(t)}),
\end{equation}
which completes the proof. 

$\Box$
\end{proof}
\textbf{Theorem \ref{thm:2}} is proved as follows.
\begin{proof}
Add $\log p(\theta)$ to both sides of (\ref{eq:52}),
\begin{align*}
l(\theta)+ \log p(\theta) &\geq l(\theta^{(t)}) + Q(\theta\vert\theta^{(t)}) - Q(\theta^{(t)}\vert\theta^{(t)})+ \log p(\theta)\\
&\begin{aligned}= \ \ 
&l(\theta^{(t)}) + \log p(\theta^{(t)}) + Q(\theta\vert\theta^{(t)}) - Q(\theta^{(t)}\vert\theta^{(t)})\\&+ \log p(\theta)-\log p(\theta^{(t)})
\end{aligned}\\
&\geq l(\theta^{(t)}) + \log p(\theta^{(t)}),
\end{align*}
where the last line follows from the assumption made in the Theorem. 

$\Box$
\end{proof}
\section{Appendix B}
\label{sec:appendixB}

\begin{longtable}{||l|c||}\caption{Local model parameters for the target state ${x}(t)$ of the MSD system. In models 2 and 3, as we adopt ARD, the bias term is pruned out; therefore, the covariance matrix is in $\mathbb{R}^{2\times 2}$.}
\label{tab:MSD_models1} \\

\hline \multicolumn{1}{|c}{\cellcolor{Gray}\textbf{ }} & \multicolumn{1}{c|}{\cellcolor{Gray}\textbf{Learned value}}\\ \hline 
\endfirsthead

\multicolumn{2}{c}%
{{\bfseries \tablename\ \thetable{} -- continued from previous page}} \\ \hline
\endhead


\hline \hline
\endlastfoot
 \hline\hline\hline
 & \\
 $\mathbf{c}_1$ & $\left(\begin{matrix} 0.16434\\-0.58072 \end{matrix}\right)$\\ 
 & \\
 $\lambda_1$ & $\left(\begin{matrix} 0.33351\\ 0.27799\end{matrix}\right)$\\
 & \\
 $\mathbf{w}_1$  & $\left(\begin{matrix} -0.2818\\
 0.87123\\0.08370 \end{matrix}\right)$\\
 & \\
 $\boldsymbol{\Sigma}_1$  & \begin{small}
 $\left(\begin{matrix} 1.46*10^{-5} & 4.22*10^{-6} &-1.43*10^{-6}\\
  4.22*10^{-6} & 5.28*10^{-5} & 4.38e*10^{-6}\\ -1.43*10^{-6} & 4.38*10^{-6} & 1.28*10^{-6} \end{matrix}\right)$
 \end{small} \\
 & \\
 $\boldsymbol{\alpha}_{1}$ & $\left(\begin{matrix} 12.59\\   1.32\\ 142.69\end{matrix}\right)$\\
 & \\
 $\beta_{f_1}$& 14281.53 \\
 & \\
 \hline\hline\hline
 & \\
 $\mathbf{c}_2$ & $\left(\begin{matrix} -0.02749\\ -0.06403 \end{matrix}\right)$ \\
 & \\
 $\lambda_2$ & $\left(\begin{matrix} 0.56134\\ 0.31700 \end{matrix}\right)$ \\
 & \\
 $\mathbf{w}_2$  & $\left(\begin{matrix} -2.69628\\8.88278 \\0\end{matrix}\right)$\\
 & \\
 $\boldsymbol{\Sigma}_2$  & \begin{small}$\left(\begin{matrix} 5.07*10^{-6} &-9.40*10^{-7}\\ -9.40*10^{-7} & 1.67*10^{-5} \end{matrix}\right)$
 \end{small} \\
 & \\
 $\boldsymbol{\alpha}_{2}$ & $\left(\begin{matrix} 13.75\\ 1.27\end{matrix}\right)$\\ 
 & \\
 $\beta_{f_2}$& 14250.26 \\
 & \\
 \hline\hline\hline
 & \\
 $\mathbf{c}_3$ & $\left(\begin{matrix} -0.07787 \\ 0.35885 \end{matrix}\right)$\\
 & \\
 $\lambda_3$ &  $\left(\begin{matrix} 0.33493\\0.25643 \end{matrix}\right)$\\
 & \\
 $\mathbf{w}_3$  & $\left(\begin{matrix} -0.05474\\0.32631\\0 \end{matrix}\right)$\\
 & \\
 $\boldsymbol{\Sigma}_3$  & \begin{small}$\left(\begin{matrix} 1.03*10^{-5} & 2.70*10^{-5} \\ 2.70*10^{-5}& 1.12*10^{-4} \end{matrix}\right)$
 \end{small} \\
 & \\
 $\boldsymbol{\alpha}_{3}$ &  $\left(\begin{matrix} 332.40\\    9.38\end{matrix}\right)$ \\ 
 & \\
 $\beta_{f_3}$& 14371.72 \\
 & \\
  \hline
\end{longtable}

\newpage
\begin{longtable}{||l|c||}\caption{Local model parameters for the response state $\dot{x}(t)$ of the MSD system. In model 2, as we adopt ARD, the bias term is pruned out; therefore, the covariance matrix is in $\mathbb{R}^{2\times 2}$.}
\label{tab:MSD_models2} \\

\hline \multicolumn{1}{|c}{\cellcolor{Gray}\textbf{ }} & \multicolumn{1}{c|}{\cellcolor{Gray}\textbf{Learned value}}\\ \hline 
\endfirsthead

\multicolumn{2}{c}%
{{\bfseries \tablename\ \thetable{} -- continued from previous page}} \\ \hline
\endhead

\hline \multicolumn{2}{|r|}{{Continued on next page}} \\ \hline
\endfoot

\hline \hline
\endlastfoot
 \hline\hline\hline
 & \\
 $\mathbf{c}_1$ & $\left(\begin{matrix} 0.16434\\-0.58072 \end{matrix}\right)$\\
 & \\
 $\lambda_1$ & $\left(\begin{matrix} 0.32576\\  0.28048\end{matrix}\right)$\\
 & \\
 $\mathbf{w}_1$  & $\left(\begin{matrix} 0.96775\\ 0.30136\\-0.55395 \end{matrix}\right)$\\
 & \\
 $\boldsymbol{\Sigma}_1$  & \begin{small}$\left(\begin{matrix} 2.11*10^{-4} & 6.38*10^{-5} & -2.10*10^{-5}\\
  6.38*10^{-5} & 7.81*10^{-4} & 6.40e*10^{-5}\\ -2.10*10^{-5} & 6.40e*10^{-5}& 1.88*10^{-5} \end{matrix}\right)$
 \end{small} \\
 & \\
 $\boldsymbol{\alpha}_{1}$ & $\left(\begin{matrix} 1.07\\   10.92\\ 3.26\end{matrix}\right)$ \\
 & \\
 $\beta_{f_1}$& 974.10 \\
 & \\
 \hline\hline\hline
 & \\
 $\mathbf{c}_2$ & $\left(\begin{matrix} -0.02749\\ -0.06403 \end{matrix}\right)$  \\
 & \\
 $\lambda_2$ &  $\left(\begin{matrix} 0.31075 \\ 0.32949 \end{matrix}\right)$\\
 & \\
 $\mathbf{w}_2$  & $\left(\begin{matrix} 0.52709\\-0.28383 \\ 0\end{matrix}\right)$\\
 & \\
 $\boldsymbol{\Sigma}_2$  & \begin{small}$\left(\begin{matrix} 7.51*10^{-5} &-2.70*10^{-5}\\ -2.70*10^{-5} & 3.32*10^{-4} \end{matrix}\right)$
 \end{small} \\
 & \\
 $\boldsymbol{\alpha}_{2}$ & $\left(\begin{matrix} 3.60\\ 12.36\end{matrix}\right)$\\ 
 & \\
 $\beta_{f_2}$& 983.51\\
 & \\
 \hline\hline\hline
 & \\
 $\mathbf{c}_3$ & $\left(\begin{matrix} -0.07787 \\ 0.35885 \end{matrix}\right)$\\
 & \\
 $\lambda_3$ &  $\left(\begin{matrix} 0.30229\\ 0.26645 \end{matrix}\right)$\\
 & \\
 $\mathbf{w}_3$  & $\left(\begin{matrix} 0.79691\\-0.25168\\ 0.24750\end{matrix}\right)$\\
 & \\
 $\boldsymbol{\Sigma}_3$  & \begin{small}$\left(\begin{matrix} 2.60*10^{-4} & 2.73*10^{-4} & 4.56*10^{-5} \\ 2.73*10^{-4}& 1.81*10^{-3} & -4.95*10^{-5}\\ 4.56*10^{-5} & -4.95*10^{-5} & 1.80*10^{-5}\end{matrix}\right)$
 \end{small} \\
 & \\
 $\boldsymbol{\alpha}_{3}$ & $\left(\begin{matrix} 1.57\\ 5.34\\16.32\end{matrix}\right)$ \\ 
 & \\
 $\beta_{f_3}$&  976.02\\
 & \\
  \hline
\end{longtable}
\newpage 
\begin{longtable}{||l|c||}\caption{Local model parameters for the response state $\theta_0$ of the SDIP system. The bias term is pruned out in models 1 and 2, and both the bias and the first linear feature are pruned in model 3.}
\label{tab:SDIP_models} \\

\hline \multicolumn{1}{|c}{\cellcolor{Gray}\textbf{ }} & \multicolumn{1}{c|}{\cellcolor{Gray}\textbf{Learned value}}\\ \hline \hline
\endfirsthead

\multicolumn{2}{c}%
{{\bfseries \tablename\ \thetable{} -- continued from previous page}} \\ \hline\hline
\endhead

\hline \multicolumn{2}{|r|}{{Continued on next page}} \\ \hline
\endfoot

\hline \hline
\endlastfoot
\hline\hline\hline
 $\mathbf{c}_1$ & $\left(\begin{matrix} 2.584*10^{-2}\\  7.47*10^{-4}\\ -5.01*10^{-3}\\ -3.64*10^{-3}\\
 -1.32*10^{-3}\\ -5.13*10^{-3}\\  1.55 \end{matrix}\right)$\\
 & \\
 $\lambda_1$ & $\left(\begin{matrix} 0.41\\  0.30\\  0.27\\  0.28\\  0.29\\  0.27\\ 12.27\end{matrix}\right)$ \\
 & \\
 $\mathbf{w}_1$  & $\left(\begin{matrix}0\\ -0.59\\ -54.63\\   0.13\\  19.07\\  12.26\\  17.96 \end{matrix}\right)$\\
 & \\
 $\boldsymbol{\Sigma}_1$  & \begin{small}
 $\left(\begin{matrix} 6.00*10^{-7}&  5.90*10^{-6}& -6.70*10^{-6}&  1.32*10^{-6}&
   1.65*10^{-7}&  3.60*10^{-6}\\
  5.90*10^{-6}&  1.52*10^{-4}& -1.03*10^{-4}&  2.02*10^{-5}&
  -7.80*10^{-6}&  5.02*10^{-5}\\
 -6.70*10^{-6}& -1.03*10^{-4}&  3.43*10^{-4}& -5.95*10^{-5}&
   1.01*10^{-5}& -2.78*10^{-4}\\
  1.32*10^{-6}&  2.02*10^{-5}& -5.95*10^{-5}& 7.28*10^{-5}&
   5.51*10^{-5}& -1.39*10^{-5}\\
  1.65*10^{-7}& -7.80*10^{-6}&  1.01*10^{-5} & 5.51*10^{-5}&
   1.50*10^{-5}& -9.23*10^{-5}\\
  3.60*10^{-6}&  5.02*10^{-5}& -2.78*10^{-4}& -1.39*10^{-5}&
  -9.24*10^{-5}&  3.09*10^{-4}\end{matrix}\right)$\end{small} \\
  & \\
 $\boldsymbol{\alpha}_{1}$ & $\left(\begin{matrix} 2.84 \\3.35*10^{-4}\\ 6.22*10^{1}\\ 2.75*10^{-3}\\
  6.65*10^{-4}\\3.10*10^{-3}\end{matrix}\right)$\\
  & \\
 $\beta_{f_1}$& 622729.59 \\
 & \\
\hline\hline\hline
 $\mathbf{c}_2$ & $\left(\begin{matrix} 0.050\\  0.003\\  0.005\\  0.003\\  0.001\\  0.003\\
 -0.102 \end{matrix}\right)$\\
 & \\
 $\lambda_2$ & $\left(\begin{matrix} 0.47\\  0.30\\   0.29\\  0.30\\  0.32\\  0.29\\  10.72\end{matrix}\right)$ \\
 & \\
 $\mathbf{w}_2$  & $\left(\begin{matrix}0\\ 1.65\\ 31.83\\  60.51\\ -7.44\\ -2.166\\ -69.39 \end{matrix}\right)$\\
 & \\
 $\boldsymbol{\Sigma}_2$  & \begin{small}
 $\left(\begin{matrix} 5.66*10^{-7}&  6.32*10^{-6}& -8.49*10^{-6}&  1.16*10^{-6}&  2.14*10^{-7}&  4.77*10^{-6}\\ 
 6.32*10^{-6} & 1.89*10^{-4}& -1.65*10^{-4} & 1.69*10^{-5}&    2.91*10^{-6} & 7.49*10^{-5}\\
   -8.49*10^{-6}& -1.65*10^{-4} & 2.65*10^{-4} &-5.83*10^{-5}&
  -1.75*10^{-5}& -1.67*10^{-4}\\1.16*10^{-6}6 & 1.69*10^{-5}& -5.83*10^{-5} & 6.79*10^{-5}&  5.04*10^{-5} &-9.07*10^{-6}\\ 2.14*10^{-7}&  2.91*10^{-6}&-1.75*10^{-5} & 5.04*10^{-5}&
   1.41*10^{-4} &-7.02*10^{-5}\\4.77*10^{-6}& 7.49*10^{-5}& -1.67*10^{-4}& -9.07*10^{-6}& -7.02*10^{-5} & 1.91*10^{-4} \end{matrix}\right)$\end{small} \\
   & \\
 $\boldsymbol{\alpha}_{2}$ & $\left(\begin{matrix}3.68*10^{-1}\\ 9.87*10^{-4}\\ 
 2.73*10^{-4}\\ 1.81*10^{-2}\\
  2.13*10^{-1}\\ 2.08*10^{-4}\end{matrix}\right)$\\ 
  & \\
 $\beta_{f_2}$& 661996.37 \\
 & \\
 \hline\hline\hline
 $\mathbf{c}_3$ & $\left(\begin{matrix} 0.021\\ 0.004\\ 0.007\\ 0.003\\ 0.003\\ 0.007\\ 0.757\end{matrix}\right)$\\
 & \\
 $\lambda_3$ & $\left(\begin{matrix} 0.27\\ 0.30\\ 0.30\\ 0.30\\ 0.289\\ 0.31\\ 6.67\end{matrix}\right)$ \\
 & \\
 $\mathbf{w}_3$  & $\left(\begin{matrix}0\\ 0 \\ 21.99\\ -60.59\\ -11.75\\ -10.16\\ 51.15 \end{matrix}\right)$\\
 & \\
 $\boldsymbol{\Sigma}_3$  & \begin{small}
 $\left(\begin{matrix}  1.22*10^{-4}& -5.85*10^{-5}& 4.884*10^{-6}& 3.230*10^{-6}& 7.690*10^{-6}\\
   -5.852*10^{-5}5& 2.67*10^{-4}& -4.86*10^{-5}& -6.793*10^{-6}& -2.04*10^{-4}\\
  4.884*10^{-6}& -4.859*10^{-5}& 5.42*10^{-5}& 3.59*10^{-5}& 1.000*10^{-5}\\
   3.230*10^{-6}& -6.793*10^{-6}& 3.59*10^{-5}5& 1.28*10^{-4}& -5.357*10^{-5}\\
  7.69*10^{-4}& -2.04*10^{-4}& 1.000*10^{-5}& -5.357*10^{-5}& 2.01*10^{-4}\end{matrix}\right)$\end{small} \\
  & \\
 $\boldsymbol{\alpha}_{3}$ & $\left(\begin{matrix} 0.002\\ 0.000\\ 0.007\\ 0.010\\ 0.000\end{matrix}\right)$\\ 
 & \\
 $\beta_{f_3}$& 657070.27 \\
 & \\
  \hline
\end{longtable}





\bibliographystyle{spmpsci}      
\bibliography{References.bib}   

%
%

\end{document}